\documentclass[journal]{IEEEtran}

\usepackage{amsmath}
\usepackage{bm}
\usepackage{color}
\usepackage{multirow}
\usepackage{graphicx}
\usepackage[round,numbers]{natbib}
\usepackage{amssymb}
\usepackage{amsthm}  
\usepackage{soul}
\setcitestyle{square}

\DeclareMathOperator*{\argmin}{argmin}
\newtheorem{remark}{Remark}

\begin{document}
\title{Cooperative Localisation of {\color{black}a} GPS-Denied {\color{black}UAV} using Direction-of-Arrival Measurements}

\author{
\IEEEauthorblockN{James S. Russell, Mengbin Ye, Brian D.O. Anderson, Hatem Hmam, Peter Sarunic}

\thanks{The work of Russell, Ye, and Anderson was supported by the Australian Research Council (ARC) Discovery Project \mbox{DP-160104500}, by 111-Project No. D17019, and by Data61-CSIRO. Ye was supported by an Australian Government Research Training Program (RTP) Scholarship. 

J.S. Russell, M.~Ye and B.D.O. Anderson are with the Research School of Engineering, Australian National University, Canberra, Australia \{\texttt{u5542624, mengbin.ye, brian.anderson\}@anu.edu.au}. B.D.O.~Anderson is also with Hangzhou Dianzi University, Hangzhou, China, and with Data61-CSIRO (formerly NICTA Ltd.) in Canberra, Australia. H. Hmam and P. Sarunic are with Australian Defence Science and Technology Group (DST Group), Edinburgh, Australia \{\texttt{hatem.hmam, peter.sarunic\}@dst.defence.gov.au}. }

}

\maketitle

\begin{abstract}
A GPS-denied UAV (Agent B) is localised {\color{black}through INS alignment} with the aid of a nearby GPS-equipped UAV (Agent A), {\color{black}which} broadcasts its position {\color{black}at several time instants}. Agent B measures the signals' direction of arrival with respect to {\color{black}Agent B's} inertial navigation frame. Semidefinite programming and the Orthogonal Procrustes algorithm are employed, and accuracy is improved through maximum likelihood estimation. The method is validated using flight data and simulations. A three-agent extension is explored.
\end{abstract}


\section{Introduction}

Unmanned aerial vehicles (UAVs) play a central role in many defence reconnaissance and surveillance operations. Formations of UAVs can provide greater reliability and coverage when compared to a single UAV. To provide meaningful data in such operations, all UAVs in a formation must have a common reference frame (typically the global frame). Traditionally, UAVs have access to the global frame via GPS. However, GPS signals may be lost in urban environments and enemy controlled airspace (jamming). {\color{black}Overcoming loss of GPS signal is a hot topic in research \citep{balamurugan2016survey}, and offers a range of different problems in literature \citep{bai2016relative,morales2016signals}.}

Without access to global coordinates, a UAV must rely on its inertial navigation system (INS). Stochastic error in on-board sensor measurements causes the INS frame to accumulate drift. At any given time, drift can be characterised by a rotation and translation with respect to the global frame, and is assumed to be independent between UAVs in a formation. As a result, INS frame drift cannot be modelled deterministically. Information from both the global and INS frames must be collected and used to determine the drift between frames {\color{black}in order to achieve INS frame alignment}. We describe this process as cooperative localisation when multiple vehicles interact to achieve the task.

Various measurement types such as distance between agents and direction of arrival of a signal (we henceforth call DOA\footnote{\color{black}A bearing generally describes a scalar measurement between two points in a plane, whereas a direction-of-arrival is a vector measurement between two points in three-dimensional ambient space, which is the space that this paper considers.}) can be used for this process. In the context of UAVs, additional sensors add weight and consume power. As a result, one generally aims to minimise the number of measurement types required for localisation. This paper studies a cooperative approach to localisation using DOA measurements.

{\color{black}When two or more GPS-enabled UAVs can simultaneously measure directions \textit{with respect to the global frame} towards the GPS-denied UAV, the location of the GPS-denied UAV is given by the point minimising distances to the half-line loci derived from the directional measurements \citep{Bishop2007, 570703, Tekdas2010}. Operational requirements may limit the number of nearby GPS-enabled UAVs to one single agent. We therefore seek a solution which does not require simultaneous measurements to a single point.

When the GPS-denied agent is able to simultaneously measure directions \textit{with respect to its local INS frame} towards multiple landmarks (two in two-dimensional space, or three in three-dimensional space) with known global coordinates, triangulation-based measurements can be used to achieve localisation. 
This problem is studied in three-dimensional space in \citep{7376225}, and in two-dimensional space in \citep{6965783, Duan2012}. 
If only one landmark bearing can be measured at any given time, a bearing-only SLAM algorithm may be used to progressively construct a map of the environment on the condition each landmark is seen at least twice.
Alignment of a GPS-denied agent's INS frame could then be achieved by determining the rotation and translation between the map's coordinate frame and the global coordinate frame.
In practice, landmark locations may be unknown, or there may be no guarantee they are stationary or permanent, and hence we require a localisation algorithm which is independent of landmarks in the environment. 
Iterative filtering methods such as the Extended Kalman Filter are often required when drift is significant between updates, however in our problem context the drift is sufficiently slow to be modelled as stationary over short periods. Given that the UAVs have limited computational capacity on board, we are motivated to formulate a localisation algorithm which does not involve an iterative filtering technique.

Without reliance on landmarks, the only directional measurements available are between the GPS-denied and the GPS-enabled UAVs. Given their potentially large separation, these UAVs are modelled as point agents, and therefore one single directional measurement is available at any given time. A stationary target is localised by an agent using bearing-only measurements in two-dimensional space \citep{Bayram2016, 256302}, and in three-dimensional space \citep{4667717}. A similar problem is considered in \cite{8351953}, in which a mobile source is localised using measurements received at a stationary receiver using an iterative filtering technique. In these works, either the entity being localised or a receiver must be stationary, and the entity requiring localisation must broadcast a signal. For operational reasons, the agent requiring localisation may be unable to broadcast signals, or agents involved may not be allowed to remain stationary. As a result, the approaches in \cite{Bayram2016}, \cite{256302}, \cite{8351953} and \cite{4667717} are not suitable. 
Commonly used computer vision techniques such as structure from motion \cite{Koenderink:91} require directional measurements towards multiple stationary points or towards a stationary point from multiple known positions. This is not possible in our problem context. The constraints flowing from the measurement and motion requirements we are imposing therefore represent a significant technical challenge. 

%
One algorithm satisfying all the constraints above was proposed in \citep{7815320}, in which two agents perform sinusoidal motion in two-dimensional ambient space and directional measurements are used to obtain the distance between Agents A and B, but localisation of B in the global frame is not achieved.

{\color{black}Motivated by interest from Australia's Defence Science and Technology Group, this paper seeks to address the problem of localising} a GPS-denied UAV, which we will call Agent B, with the assistance of a GPS-enabled UAV, which we will call Agent A. Both agents move arbitrarily in three-dimensional space. {\color{black}Agent B navigates using an INS frame.} Agent A broadcasts its position in the global coordinate frame at discrete instants in time. For each broadcast of Agent A, Agent B is able to take a DOA measurement towards Agent A. 

The problem setup and the solution we propose are both novel. In particular, while the literature discussed above considers certain aspects from the following list, none consider all of the following aspects simultaneously:
\begin{itemize}
\item The network consists of only two mobile agents (and is therefore different to the sensor network localisation problems in the literature).
\item There is no a priori knowledge or sensing of a stationary reference point in the global frame.
\item The UAVs execute unconstrained arbitrary motion in three-dimensional space.
\item Cooperation\footnote{\color{black}Agent A's role in the cooperation is to broadcast its position over a series of time instants.} occurs between a GPS-enabled and a GPS-denied UAV (transmission of signals is unidirectional; from Agent A to Agent B).
\end{itemize}

It is the combination of all the above aspects which make the problem significantly more difficult and thus existing methods are unsuitable. In \citep{7798924}, this problem is studied in two-dimensional space using bearing measurements. One single piece of data is acquired at each time step. In this paper, each DOA measurement gives two scalar quantities, and adding the third dimension significantly complicates the problem, thus requiring new techniques to be introduced.

{\color{black} 
In our proposed solution, we localise Agent B by identifying the relationship between the global frame (navigated by Agent A) and the inertial navigation frame of Agent B. 
The relationship is identified by solving a system of linear equations for a set of unknown variables. The nature of the problem means quadratic constraints exist on some of the variables. To improve robustness against noisy measurements, we exploit the quadratic constraints and use semidefinite programming (SDP) and the Orthogonal Procrustes algorithm to obtain an initial solution for maximum likelihood (ML) estimation. Our approach combining SDP, the Orthogonal Procrustes algorithm and ML estimation is a key novel contribution over existing works.

We evaluate the performance of the algorithm by (i) using a real set of trajectories and (ii) using Monte Carlo simulations. Sets of unsuitable trajectories are identified, in which our proposed method cannot feasibly obtain a unique solution. Finally, we explore an extension of the algorithm to a three-agent network in which two agents are GPS-denied.

The rest of the paper is structured as follows. In Section~\ref{sec:problem_def} the problem is formalised. In Section~\ref{sec:noiseless} a localisation method using a linear equation formulation is proposed. Section~\ref{sec:noisy_measurements} extends this method to semidefinite programming to produce a more robust localisation algorithm. 
In Section \ref{sec:mle}, a maximum likelihood estimation method is presented to refine results further.
Section~\ref{sec:simulations} presents simulation results to evaluate the performance of the combined localisation algorithm. 
Section VII extends the localisation algorithm to a three-agent network. 
The paper is concluded in Section~\ref{sec:conclusion}. \footnote{Early sections in this paper (covering up to and including employment of Orthogonal Procrustes algorithm) appeared in less detail in the conference paper \citep{RUSSELL20178019}. Additions have been made to these sections - the literature review is now more extensive; the role of different coordinate frames is much more explicitly set out; the algorithm's performance is now validated on real flight trajectories, and the inclusion of apparently redundant quadratic constraints in the SDP problem formulation is justified further. Analysis of unsuitable trajectories, ML refinement and the three-agent extension are further extensions beyond \cite{RUSSELL20178019}.}

\section{Problem Definition}\label{sec:problem_def}

{\color{black}

{\color{black}

Two agents, which we call Agent A and Agent B, travel along arbitrary trajectories in three-dimensional space. Agent A has GPS and therefore navigates with respect to the global frame. Because Agent B cannot access GPS, it has no ability to self-localise in the global frame, but can self-localise and navigate in a local inertial frame by integrating gyroscope and accelerometer measurements. 

This two-agent localisation problem involves 4 frames as in Figure \ref{fig:frames}. The importance of each frame, and its use in obtaining the localisation, will be made clear in the sequel.
\begin{figure}[b]
\includegraphics[width=\linewidth]{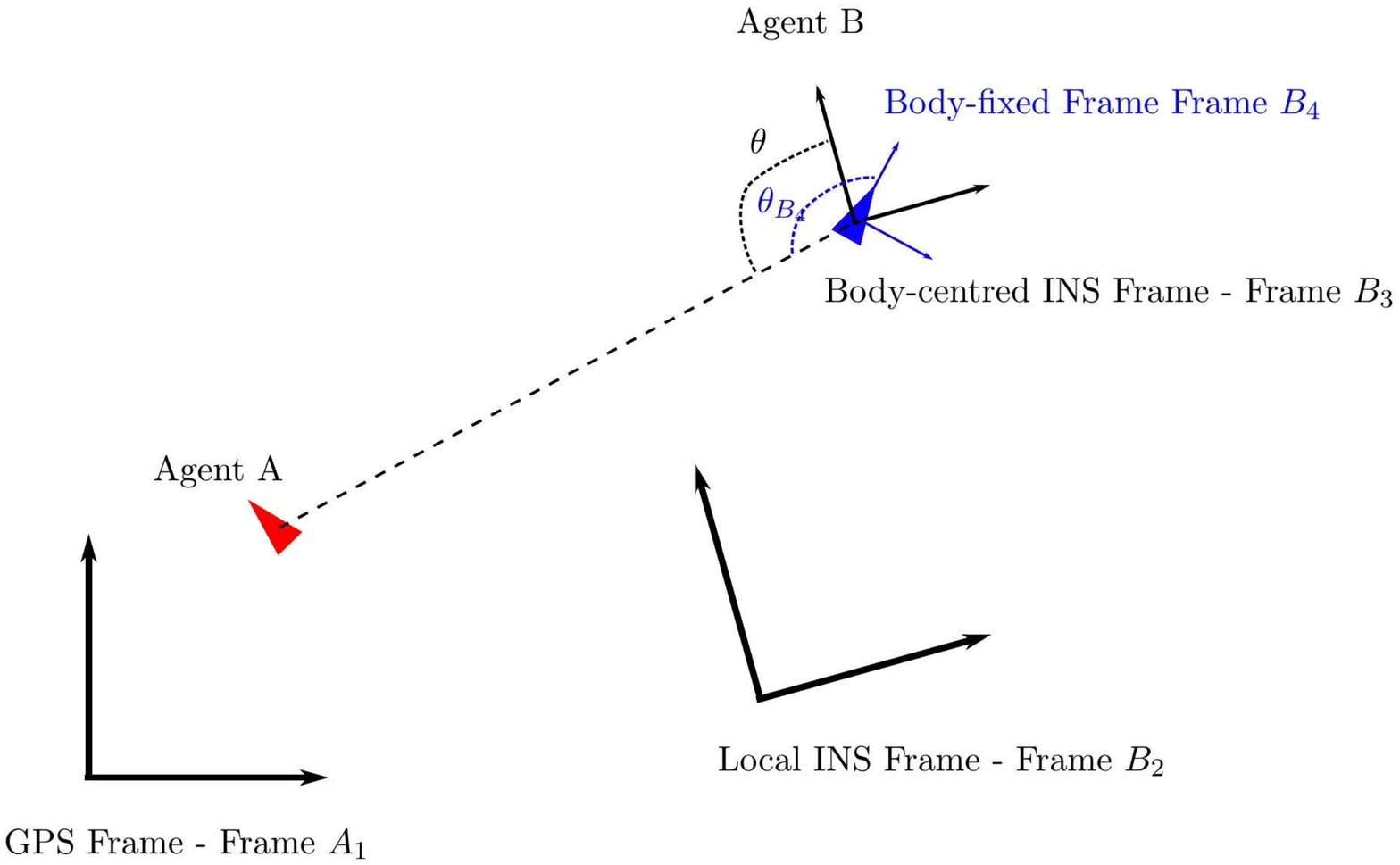}
\caption{Illustration of coordinate frames in a two-dimensional space}
\label{fig:frames}
\end{figure}
Frames are labelled as follows:
\begin{itemize}
\item Let $A_1$ denote the global frame (available only to Agent A), 
\item let $B_2$ denote the local INS frame of Agent B, 
\item let $B_3$ denote the body-centred INS frame of Agent B (axes of frames $B_2$ and $B_3$ are parallel by definition),
\item let $B_4$ denote the body-fixed frame of Agent B.
\end{itemize}

{\color{black}The expression of directional measurements with respect to the INS frame in vector form motivates the definition of the body-centred frame $B_3$. Later, we find that differences in body fixed frame azimuth and elevation measurement noise motivate the use of $B_4$ when discussing maximum likelihood estimation.}

Note that agents A and B are denoted by a single letter, whereas frames $A_1$ and $B_i$ for $i$ = 2,3,4 are denoted by a letter-number pair. Positions of each agent in their respective navigation frames ($A_1$ and $B_2$)
are obtained through a discrete-time measurement process. 
Let {\color{black}$\bm{p^{I_0}_J}(k)$} denote the position of Agent $J$ in coordinates of frame $I_0$ at the $k^{th}$ time instant. Let $u_J$, $v_J$, $w_J$ denote {\color{black}Agent J's} coordinates in the global frame ($A_1$), and $x_J$, $y_J$, $z_J$ denote {\color{black}Agent J's} coordinates in Agent B's local INS frame ($B_2$).} It follows that:
\begin{align}
\bm{p_A^{A_1}}(k) = [u_A(k), \; v_A(k), \; w_A(k)] ^\top \\
\bm{p_B^{B_2}}(k) = [x_B(k), \; y_B(k), \; z_B(k)] ^\top
\end{align}
The rotation and translation of Agent B's local INS frame ($B_2$) with respect to the global frame ($A_1$) evolves via drift. Although this drift is significant over long periods, frame $B_2$ can be modelled as stationary with respect to frame $A_1$ over short intervals\footnote{\color{black}If loss of GPS is sustained for extensive periods we recommend using the algorithm in this paper as an initialisation for a recursive filtering algorithm.}. During these short intervals, the following measurement process occurs multiple times. At each time instant $k$, the following four activities occur simultaneously\footnote{\color{black}The relative speed of the UAVs with respect to the speed of light is negligible.}:
\begin{itemize}
\item Agent B records its own position in the INS frame $\bm{p_B^{B_2}}(k)$.
\item Agent A records and broadcasts its position in the global frame $\bm{p_A^{A_1}}(k)$.
\item Agent B receives the broadcast of $\bm{p_A^{A_1}}(k)$ from Agent A, and measures this signal's DOA using instruments fixed to the UAV's fuselage. This directional measurement is therefore naturally referenced to the body-fixed frame $B_4$. 
\item {\color{black}Agent B's attitude, i.e. orientation with respect to the INS frames $B_2$ and $B_3$  is known. An expression for the DOA measurement referenced to the axes INS frames $B_3$ can therefore be easily calculated.}
\end{itemize}

A DOA measurement, {\color{black}referenced to a frame with axes denoted $x, y, z$, is expressed as follows:}
\begin{itemize}
\item Azimuth ($\theta$): angle formed between the positive $x$ axis and the projection of the {\color{black}free vector} from Agent B towards Agent A onto the $xy$ plane.
\item Elevation ($\phi$): angle formed between the {\color{black}free vector from Agent B towards Agent A} and $xy$ plane. The angle is positive if the $z$ component of the unit vector towards Agent A is positive.
\end{itemize}




An illustration of azimuth and elevation components of a DOA measurement is provided in Figure \ref{fig:doa_diagram}.

\begin{figure}[b]
	\includegraphics[width=0.8\linewidth]{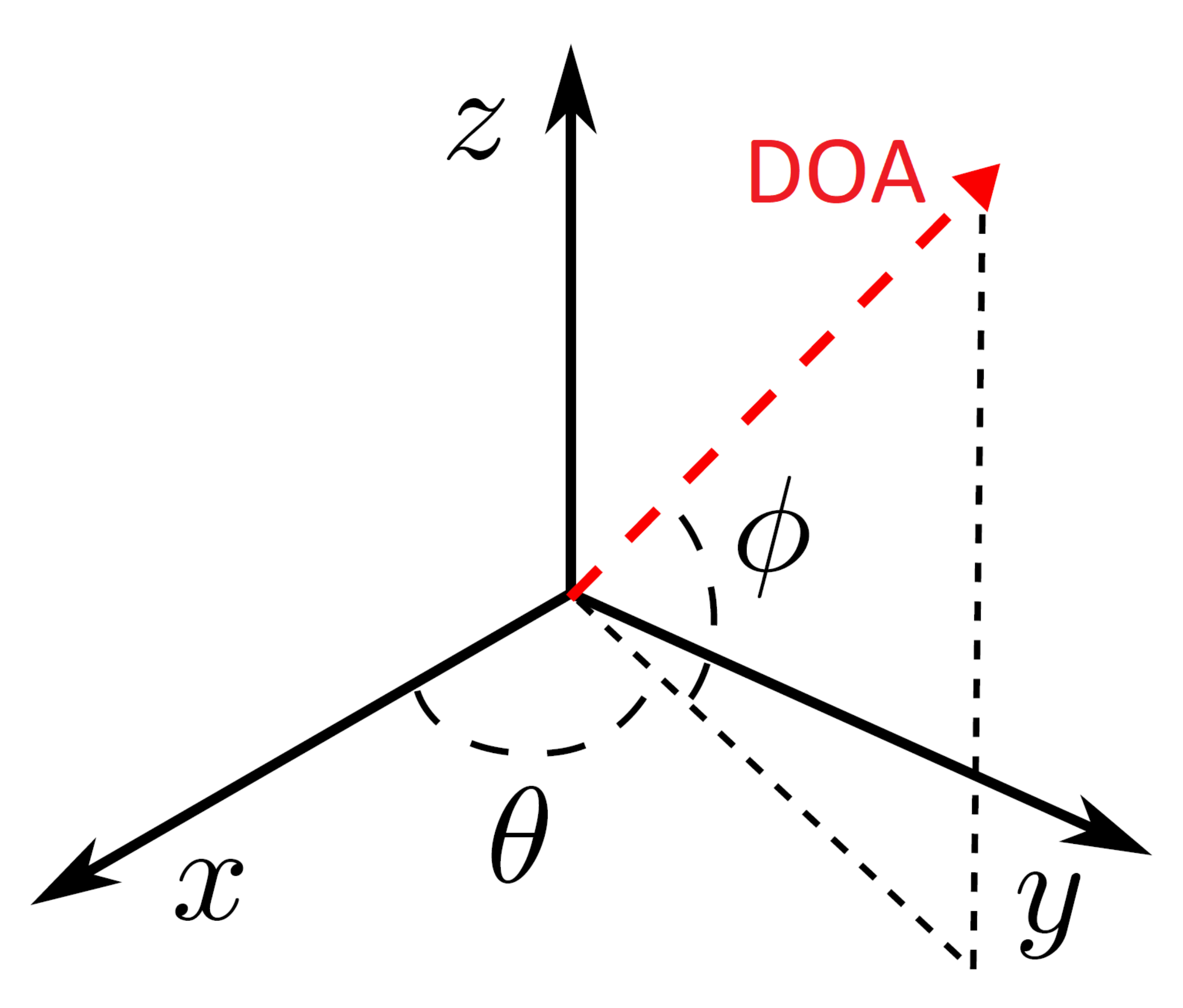}
	\caption{Illustration of azimuth and elevation components of a DOA measurement}
	\label{fig:doa_diagram}
\end{figure}

{\color{black} The problem addressed in this paper, namely the localisation of Agent B,  is achieved if we can determine the relationship between the global frame $A_1$ and the local INS frame $B_2$. This information can be used to determine global coordinates of Agent B at each time instant $k$}:
\begin{equation}
\bm{p_B^{A_1}}(k) = [u_B(k), \; v_B(k), \; w_B(k)] ^\top 
\end{equation}
Passing between the global frame ($A_1$) and the local INS frame of Agent B ($B_2$) is achieved by a rotation of frame axes (defined by a rotation matrix, call it $\bm{R_{A_1}^{B_2}}$) and translation $\bm{t_{A_1}^{B_2}}$ of frame. For instance, the coordinate vector of the position of Agent A referenced to the INS frame of Agent B is:
{\color{black}
\begin{equation}
\label{eq:trans}
\bm{p_A^{B_2}}(k) = \bm{R^{B_2}_{A_1}} \, \bm{p_A^{A_1}}(k) + \bm{t_{A_1}^{B_2}}
\end{equation}
}
We therefore have 
\begin{equation}
\label{eq:inverse_trans}
\bm{p_B^{A_1}}(k) = \bm{R^{B_2\top}_{A_1}} \, (\bm{p_B^{B_2}}(k) - \bm{t_{A_1}^{B_2}})
\end{equation}
where $\bm{R^{B_2\top}_{A_1}} = \bm{R^{A_1}_{B_2}} $ and $-\bm{R^{B_2\top}_{A_1}} \, \bm{t_{A_1}^{B_2}} = \bm{t_{B_2}^{A2}}$. The localisation problem can be reduced to solving for $\bm{R_{A_1}^{B_2}} \in SO(3)$ with entries $r_{ij}$ and $\bm{t_{A_1}^{B_2}} \in \mathbb{R}^3$ with entries $t_i$.



{\color{black} The matrix $\bm{R_{A_1}^{B_2}}$ is a rotation matrix if and only if $\bm{R_{A_1}^{B_2}}\bm{R_{A_1}^{B_2\top}} = I_3$ and $\det(\bm{R_{A_1}^{B_2}}) = 1$. As will be seen in the sequel, these constraints are equivalent to a set of quadratic constraints on the entries of $\bm{R_{A_1}^{B_2}}$. In total there are 12 entries of $\bm{R_{A_1}^{B_2}}$ and $\bm{t_{A_1}^{B_2}}$ to be found as we work directly with $r_{ij}$.}

\section{Linear System Method}\label{sec:noiseless}

{\color{black}This section presents a linear system (LS) method to solving the localisation problem. Given enough measurements, the linear system approach can achieve exact localisation when using noiseless DOA measurements, so long as Agents A and B avoid a set of unsuitable trajectories (which are detailed in Section VI-C) in which rank-deficiency is encountered. Building on this, Section \ref{sec:noisy_measurements} introduces non-linear constraints to the linear problem defined in this section to improve accuracy in the presence of noise.}

\subsection{Forming a system of linear equations}

The following analysis holds for all $k$ instants in time, hence we drop the argument $k$. {\color{black} The DOA measurement can be represented by a unit vector pointing from Agent B to Agent A. This vector is defined by azimuth and elevation angles $\theta$ and $\phi$ referenced to the local INS frame $B_2$, and its coordinates in the frame $B_2$ are given by:}
\begin{equation}
\bm{\hat{q}}(\theta, \phi)= [\hat{q_1}, \hat{q_2}, \hat{q_3}] = [\cos\theta\cos\phi, \sin\theta\cos\phi, \sin\phi]^\top
\end{equation}
Define $\bar q \doteq \Vert \bm{p_A^{B_2}} - \bm{p_B^{B_2}}\Vert$ as the Euclidean distance between Agent A and Agent B (which is not available to either agent). Scaling to obtain the unit vector $\bm{\hat{q}}$ gives
\begin{equation}
\label{eq:system1}
\bm{\hat{q}}(\theta, \phi)
=
\frac{1}{\bar q}
\begin{bmatrix}
    x_A - x_B, &
    y_A - y_B, &
    z_A - z_B
\end{bmatrix}^\top
\end{equation}
Applying equation (\ref{eq:trans}) yields:
\begin{equation}
\label{eq:scaled}
\begin{bmatrix}
	{\color{black}\hat{q_1}} \\
    {\color{black}\hat{q_2}} \\
    {\color{black}\hat{q_3}} \\
\end{bmatrix}
=
\frac{1}{\bar q}
\begin{bmatrix}
    r_{11} u_A + r_{12} v_A + r_{13} w_A + t_1 - x_B \\
    r_{21} u_A + r_{22} v_A + r_{23} w_A + t_2 - y_B \\
    r_{31} u_A + r_{32} v_A + r_{33} w_A + t_3 - z_B \\
\end{bmatrix}
\end{equation}
The left hand vector is calculated directly from DOA measurements. Cross-multiplying entries 1 and 3 of both vectors eliminates the unknown $\bar q$, and rearranging yields:
\begin{align}
\begin{split}
\label{eq:xzlinkb}
& (u_A{\color{black}\hat{q_3}})r_{11} + (v_A{\color{black}\hat{q_3}})r_{12} + (w_A{\color{black}\hat{q_3}})r_{13}
 - (u_A{\color{black}\hat{q_1}})r_{31} \\
 & - (v_A{\color{black}\hat{q_1}})r_{32} - (w_A{\color{black}\hat{q_1}})r_{33} + ({\color{black}\hat{q_3}})t_1 - ({\color{black}\hat{q_1}})t_3 \\
& = ({\color{black}\hat{q_3}})x_B - ({\color{black}\hat{q_1}})z_B
\end{split}
\end{align}
Similarly, from the second and third entries in (\ref{eq:scaled})
\begin{align}
\begin{split}
\label{eq:yzlinkb}
& (u_A{\color{black}\hat{q_3}})r_{21} + (v_A{\color{black}\hat{q_3}})r_{22} + (w_A{\color{black}\hat{q_3}})r_{23} - (u_A{\color{black}\hat{q_2}})r_{31} \\
& - (v_A{\color{black}\hat{q_2}})r_{32} - (w_A{\color{black}\hat{q_2}})r_{33} + ({\color{black}\hat{q_3}})t_2 - ({\color{black}\hat{q_2}})t_3 \\
& = ({\color{black}\hat{q_3}})y_B - ({\color{black}\hat{q_2}})z_B
\end{split}
\end{align}
\emph{Notice that both equations 
\eqref{eq:xzlinkb} and \eqref{eq:yzlinkb} 
are linear in the unknown $r_{ij}$ and $t_i$ terms.} Given a series of $K$ DOA measurements (each giving $\phi(k), \theta(k)$), \eqref{eq:xzlinkb} and \eqref{eq:yzlinkb} can then be used to construct the following system of linear equations:
\begin{equation}
\label{eq:Axb}
\bm{A}\bm{\Psi} = \bm{b}\, , \quad \bm{A} \in \mathbb{R}^{2K \times 12}
\end{equation}
{\color{black} where $\bm{A}$, $\bm{b}$ are completely known, containing $\theta(k)$, $\phi(k)$, $\bm{p_A^{A_1}}$ and $\bm{p_B^{B_2}}$}. The 12-vector of unknowns $\bm{\Psi}$ is defined as:
\begin{equation}
\label{eq:psi}
\bm{\Psi} = [r_{11} \enspace r_{12} \enspace r_{13} \enspace ... \enspace r_{31} \enspace r_{32} \enspace r_{33} \enspace t_1 \enspace t_2\enspace t_3]^\top
\end{equation}
{\color{black}Entry-wise definitions of $\bm{A}$ and $\bm{b}$ are listed in Appendix A.
These entries of $\bm{\Psi}$ can be used to reconstruct the trajectory of Agent B in the global frame using (\ref{eq:inverse_trans}), and therefore \color{black}solving \eqref{eq:Axb} for $\bm{\Psi}$} constitutes as a solution to the localisation problem.

If $K \geq 6$, the matrix $\bm{A}$ will be square or tall. 
In the noiseless case, if $\bm{A}$ is of full column rank, equation (\ref{eq:Axb}) will be solvable.



{\color{black}\subsection{Example of LS method in noiseless case using real flight trajectories}\label{ssec:example_noiseless}}

We demonstrate the linear system method using trajectories performed by aircraft operated by the Australian Defence Science and Technology Group. Positions of both Agent A {\color{black}and B} within the global frame and Agent B within the INS frame were measured by on-board instruments, {\color{black}whereas we generated a set of \textit{calculated} DOA values using the above recorded real measurements.}

These trajectories are plotted in Figure \ref{fig:frame_with_legend}. 
We will make additional use of this trajectory pair in the noisy measurement case {\color{black}presented in Section~\ref{sec:noisy_measurements}}, and in the maximum likelihood estimation refinement of the noisy case localisation result in Section \ref{sec:mle}. Extensive Monte Carlo simulations demonstrating localisation for a large number of realistic\footnote{By realistic, we mean that the distance separation between successive measurements is consistent with UAV flight speeds and ensures the UAV does not exceed an upper bound on the turn/climb rate. Further detail is provided \color{black}in Section VI-B.} flight trajectories are left to the noisy measurement case.

The rotation matrix and translation vector describing the relationship between the global frame and Agent B's inertial navigation frame are:
\begin{equation}
\bm{R_{A_1}^{B_2}} = 
\begin{bmatrix}
1.000 & -0.032 & 3.78\times 10^{-5} \\
0.032 & 1.000 & 0.002 \\
-9.48\times 10^{-5} & -0.002 & 1.000
\end{bmatrix}
\end{equation}
\begin{equation} 
\bm{T_{A_1}^{B_2}} = 
\begin{bmatrix}
854.87 \quad 6.18 \quad 1.93 \\
\end{bmatrix} ^\top
\end{equation}  
{\color{black}Azimuth and elevation angle measurements 
are tabulated in Table \ref{tab:data}. Using \eqref{eq:Axb}, $\bm{R_{A_1}^{B_2}}$ and $\bm{T_{A_1}^{B_2}}$ were obtained exactly for the given flight trajectories; the solution is shown by the green line in Figure \ref{fig:frame_with_legend}.}

\begin{table*}
	\centering
	\caption{\color{black}Positions of Agents A and B, noiseless DOA measurements, and SDP+O+ML solution for real trajectory pair with noisy DOA measurements}
	\label{tab:data}
	\begin{tabular}{||c c c c c c||} 
		\hline
		\vspace{0.5mm}
		$k$ & $\bm{p_A^{A_1}}$ [m] & $\bm{p_B^{B_2}}$ [m] & $\bm{p_B^{A_1}}$ [m] & DOA [$\theta$ $\phi$] [rads] & $\overline{\bm{p_B^{A_1}}}$ using SDP+O+ML\\
		\hline\hline
		1 & $[349.1 \  -924.1 \  374.4]^\top$ & $[1039.2 \  574.2 \  311.3]^\top$ & $[202.5 \  561.3 \  310.4]^\top$ & [-1.4403 \  0.0447] & $[135.9 \  468.16 \  276.2]^\top$\\
		\hline
		2 & $[781.0 \  -870.3 \  372.5]^\top$ & $[1486.1 \  519.4 \  310.9]^\top$ & $[647.3 \  492.1 \  309.9]^\top$ & [-1.4409 \  0.0474] & $[583.6 \  426.3 \  297.9]^\top$\\
		\hline
		3 & $[1007.0 \  -522.7 \  373.3]^\top$ & $[1946.2 \  458.2 \  310.2]^\top$ & $[1105.2 \  416.2 \  309.1]^\top$ & [-1.6430 \  0.0697] & $[1044.8 \  378.6 \  319.9]^\top$\\
		\hline
		4 & $[869.8 \  -91.3 \  373.2]^\top$ & $[2140.4 \  746.9 \  309.8]^\top$ & $[1308.6 \  698.5 \  309.2]^\top$ & [-2.0459 \  0.0723] & $[1230.3 \  672.8 \  330.6]^\top$\\
		\hline
		5 & $[431.4 \  56.6 \  373.1]^\top$ & $[2201.6 \  1166.4 \  308.8]^\top$ & $[1383.2 \  1115.8 \  309.0]^\top$ & [-2.2708 \  0.0464] & $[1279.0 \  1093.9 \  334.7]^\top$\\
		\hline
		6 & $[33.9 \  -262.2 \  373.6]^\top$ & $[2032.8 \  1477.7 \  310.2]^\top$ & $[1224.5 \  1432.5 \  310.9]^\top$ & [-2.1512 \  0.0317] & $[1101.3 \  1400.2 \  329.2]^\top$\\
		\hline
	\end{tabular}
\end{table*}
\section{Semidefinite programming method}\label{sec:noisy_measurements}

This section presents a semidefinite programming (SDP) method for localisation, extending from the linear system (LS) approach presented in Section \ref{sec:noiseless}. This method reduces the minimum required number of DOA measurements to obtain a unique solution, and is more robust than LS in terms of DOA measurement noise {\color{black}}and unsuitable trajectories are reduced.

Rank-relaxed SDP is used to incorporate the quadratic constraints on certain entries of $\bm{\Psi}$ arising from the properties of rotation matrices. 
{\color{black}The inclusion of rotation matrix constraints in SDP problems has been used previously to jointly estimate the attitude and spin-rate of a satellite \cite{A2}, and in camera pose estimation using SFM techniques when directional measurements are made to multiple points simultaneously \cite{A3}. We now apply this technique in a novel context to achieve INS alignment of Agent B, sufficient for its localisation.} Finally, the Orthogonal Procrustes algorithm (O) is used to compensate for the rank relaxation of the SDP.
\subsection{Quadratic constraints on entries of $\bm{\Psi}$}
\label{ssec:constraints_psi}

Rank-relaxed semidefinite programming {\color{black}(in the presence of inexact or noise contaminated data)} benefits from the inclusion of quadratic constraint equations. We now identify 21 quadratic and linearly independent constraint equations on entries of $\bm{R_{A_1}^{B_2}}$, which all appear in $\bm\Psi$ in (\ref{eq:psi}). Recall the orthogonality property of rotation matrices $\bm{R_{A_1}^{B_2}}\bm{{R_{A_1}^{B_2}}^\top} = I_3$. By computing each entry of $\bm{R_{A_1}^{B_2}}\bm{{R_{A_1}^{B_2}}^\top}$, setting these equal to entries of $\bm{I_3}$, and referencing the $i^{th}$ entry of $\bm\Psi$ as $\psi_i$, we define constraints:

\begin{subequations}\label{eq:constraints_psi_01}
\begin{align}
& C_1 = \psi_1^2 + \psi_2^2 + \psi_3^2 - 1 = 0 \\
& C_2 = \psi_4^2 + \psi_5^2 + \psi_6^2 - 1 = 0 \\
& C_3 = \psi_7^2 + \psi_8^2 + \psi_9^2 - 1 = 0 \\
& C_4 = \psi_1\psi_4 + \psi_2\psi_5 + \psi_3\psi_6 = 0 \\
& C_5 = \psi_1\psi_7 + \psi_2\psi_8 + \psi_3\psi_9 = 0 \\
& C_6 = \psi_4\psi_7 + \psi_5\psi_8 + \psi_6\psi_9 = 0
\end{align}
\end{subequations}
To simplify notation we call $C_{j:k}$ the set of constraints $C_i$ for $i = j, .., k$. Similarly, by computing each entry of $\bm{{R_{A_1}^{B_2}}^\top}\bm{R_{A_1}^{B_2}}$ and setting these equal to $\bm{I_3}$, we define constraints $C_{7:12}$. We omit presentation of $C_{7:12}$ due to space limitations and similarity with $C_{1:6}$. The sets $C_{1:6}$ and $C_{7:12}$ are clearly equivalent.

Further constraints are required to ensure $\det(\bm{R_{A_1}^{B_2}}) = 1$. Cramer's formula states that
${\bm{R_{A_1}^{B_2}}}^{-1} = \text{adj}(\bm{R_{A_1}^{B_2}})/\det(\bm{R_{A_1}^{B_2}})$,
where $\text{adj}(\bm{R_{A_1}^{B_2}})$ denotes the adjugate matrix of $\bm{R_{A_1}^{B_2}}$. Orthogonality of $\bm{R_{A_1}^{B_2}}$ implies ${\bm{R_{A_1}^{B_2}}}^\top = \text{adj}(\bm{R_{A_1}^{B_2}})$ or that $\bm{R_{A_1}^{B_2}} = \text{adj}(\bm{R_{A_1}^{B_2}})^\top$. By computing entries of the first column of $\bm{Z} = \bm{R_{A_1}^{B_2}} - \text{adj}(\bm{R_{A_1}^{B_2}})^\top$ and setting these equal to $0$, we define constraints $C_{13:15}$:
\begin{subequations}\label{eq:constraints_psi_02}
\begin{align}
& C_{13} = \psi_1 - (\psi_5\psi_9 - \psi_6\psi_8) = 0 \\
& C_{14} = \psi_4 - (\psi_3\psi_8 - \psi_2\psi_9) = 0 \\
& C_{15} = \psi_7 - (\psi_2\psi_6 - \psi_3\psi_5) = 0
\end{align}
\end{subequations}
Similarly, by computing the entries of the second and third columns of $\bm{Z}$ and setting these equal to $0$, we define constraints $C_{16:18}$ and $C_{19:21}$ respectively. Due to space limitations, we omit presenting them. The complete set $C_{1:21} \doteq C_{\Psi}$ constrains $\bm{R_{A_1}^{B_2}}$ to be a rotation matrix. {\color{black}The set of constraints is not an independent set, e.g. $\bm{C_{1:6}}$ is equivalent to $\bm{C_{7:12}}$. The benefits of the inclusion of dependent constraints is discussed further in Section IV-C.}

As we will show below, due to these additional relations, localisation requires 
azimuth and elevation measurements at 
a minimum of 4 instants only ($K = 4$), as opposed to 6 instants required in Section \ref{sec:noiseless}.

\subsection{Formulation of the Semidefinite Program}\label{ssec:sdp_def}
The goal of the semidefinite program is to obtain:
\begin{equation}
\label{eq:non-sdp_objective}
\argmin_{\bm{\Psi}}||\bm{A\Psi}-\bm{b}||
\end{equation}
subject to $C_{\Psi}$. Equivalently, we seek $\argmin_\Psi ||\bm{A\Psi}-\bm{b}||^2$ subject to $C_{\Psi}$. We define the inner product of two matrices $\bm{\bm{\bm{U}}}$ and $\bm{V}$ as $\langle \bm{\bm{\bm{U}}},\bm{V} \rangle = \text{trace}(\bm{\bm{\bm{U}}}\bm{V}^\top)$. One obtains
\begin{align}
||\bm{A\Psi-b}||^2 = \left\langle \bm{P} , \bm{X} \right\rangle
\end{align}
where $\bm{P} = 
\begin{bmatrix}
    \bm{A} \quad \bm{b}
\end{bmatrix}^\top
\begin{bmatrix}
    \bm{A} \quad \bm{b}
\end{bmatrix}
$ and 
$
\bm{X} = 
[
    \bm{\Psi}^\top \; -1
]^\top
[
    \bm{\Psi}^\top \; -1
]
$
and $\bm{X}$ is a rank 1 positive-semidefinite matrix\footnote{All matrices $M$ which can be expressed in the form of $\bm{M = v^\top v}$ where $v$ is a row vector are positive-semidefinite matrices.}. The constraints $C_{\Psi}$ can also be expressed in inner product form. For $i = 1, ..., 21$, $C_i = 0$ is equivalent to $\langle \bm{Q_i}, \bm{X} \rangle = 0$ for some easily determined $\bm{Q_i}$. Solving for $\bm{\Psi}$ in (\ref{eq:non-sdp_objective}) is therefore equivalent to solving for:
\begin{align}
\argmin_\Psi& \langle \bm{P}, \bm{X} \rangle \\
\label{eq:sdpconstr1}
\bm{X} &\geq 0 \\
\label{eq:sdpconstr2}
\text{rank}(\bm{X}) &= 1 \\
\label{eq:sdpconstr3}
\bm{X_{13,13}} &= 1 \\
\langle \bm{Q_i}, \bm{X} \rangle &= 0 \quad \quad i = 1, ..., 21
\end{align}
\subsection{Rank Relaxation of Semidefinite Program}\label{ssec:relax_sdp}
 
This semidefinite program is a reformulation of a quadratically constrained quadratic program (QCQP). Computationally speaking, QCQP problems are generally NP-hard. A close approximation to the true solution can be obtained in polynomial time if the rank 1 constraint on $\bm{X}$, i.e. (22), is relaxed. {\color{black} A full technical explanation of semidefinite relaxation, and discussion on 
its applicability can be found in \cite{A1}.} This relaxation significantly increases the dimension of the SDP solver's co-domain. A notable consequence is that dependent constraints which are linearly independent over $\mathbb R$ within $\bm{C_\Psi}$, such as sets $\bm{C_{1:6}}$ and $\bm{C_{7:12}}$, cease to be redundant when expressed in inner-product form and applied to entries of $\bm{X}$. Hypothesis testing using extensive simulations confirmed with confidence above 95\% that inclusion of quadratically dependent constraints causes an improvement in localisation accuracy.

The solution to the relaxed semidefinite program $\bm{X}$ is typically close to being a rank 1 matrix\footnote{The measure used for closeness to rank 1 is the ratio of the two largest singular values in the singular value decomposition of $\bm{X}$.} when DOA measurements are noisy. The closest rank 1 approximation to $\bm{X}$, which we call $\bm{\hat{X}}$, is obtained by evaluating the singular value decomposition of $\bm{X}$, then setting all singular values except the largest equal to zero. From $\bm{\hat{X}}$, one can then use the definition of $\bm{X}$ to obtain the approximation of $\bm{\Psi}$, which we will call $\bm{\hat\Psi}$. Entries $\hat\psi_{i}$ for $i = 10, 11, 12$ can be used immediately to construct an estimate for $\bm{t_{A_1}^{B_2}}$, which we will call $\bm{\bm{\overline{t}}}$. Entries $\hat\psi_{i}$ for $i = 1, ..., 9$ will be used to construct an intermediate approximation of $\bm{R_{A_1}^{B_2}}$, which we call $\bm{\widehat{R}}$, and which we will refine further.




\subsection{Orthogonal Procrustes Algorithm}\label{ssec:procrustes}
Due to the relaxation of the rank constraint (23) on $\bm{X}$, it is no longer guaranteed that entries of $\bm{\hat\Psi}$ strictly satisfy the set of constraints $C_{\Psi}$. Specifically, the matrix $\bm{\widehat{R}}$ may not be a rotation matrix. The Orthogonal Procrustes algorithm is a commonly used tool to determine the closest orthogonal matrix (denoted $\bm{\overline{R}})$ to a given matrix, $\bm{\widehat{R}}$. This is given by $\bm{\overline{R}} = \argmin_{\bm{\Omega}} || \bm{\Omega} - \bm{\widehat{R}}||_F$, subject to $\bm{\Omega}\bm{\Omega}^\top = I$, where $||.||_F$ is the Frobenius norm. 

When noise is high, the above method occasionally returns $\bm{\overline{R}}$ such that $\det(\bm{\overline{R}}) = -1$. {\color{black}In this case, we employ a special but standard case of the Orthogonal Procrustes algorithm \cite{Eggert1997} to ensure we obtain rotation matrices and avoid reflections by flipping the last column in one of the unitary matrix factors of the singular value decomposition.}

The matrix $\bm{\overline{R}}$ and vector $\bm{\bm{\overline{t}}}$ are the final estimates of $\bm{R_{A_1}^{B_2}}$ and $\bm{t_{A_1}^{B_2}}$ using semidefinite programming and the Orthogonal Procrustes algorithm. The estimate of Agent B's position in the global frame is $\bm{\overline{p_B^{A_1}}} = \bm{\overline{R}}^\top(\bm{p_B^{B_2}} - \bm{\bm{\overline{t}}})$.

{\color{black}For convenience, we use SDP+O to refer to} the process of solving a rank-relaxed semidefinite program, and then applying the Orthogonal Procrustes algorithm to the result.

\subsection{Example of SDP+O method with noisy DOA measurements} \label{ssec:example_noisy}

In this subsection, we apply the SDP+O method to perform localisation in a noisy DOA measurement case using the real trajectory example from Section \ref{sec:noiseless}.
{\color{black}A popular practice for performing DOA measurements from Agent B towards Agent A is to use fixed RF-antennas and/or optical sensors on board Agent B's airframe.} The horizontal RF antenna typically has a larger aperture (generally around 4 times) than the vertical RF antenna. This is due to the typical ratio of a fixed-wing UAV's wingspan to its fuselage height. As a result, errors in azimuth and elevation measurements, referenced to the body-fixed frame $B_4$, are modelled by independent zero-mean Gaussian distributed variables with different standard deviations, denoted $\sigma_{\Theta}$ and $\sigma_{\Phi}$.

We now add noise (for the purposes of this simulation example based on the real data) to the calculations of body-fixed frame azimuth and elevation components of DOA. Strictly speaking, each noise is expected to follow a von Mises distribution, {\color{black}which generalises a Gaussian distribution to a circle \citep{forbes2011statistical_book}}. For small noise, as often encountered, the von Mises distribution can be approximated by a Gaussian distribution. In this example we assume {\color{black}body-fixed frame} azimuth and elevation measurement errors have standard deviations of $\sigma_\Theta = 0.5^\circ$ and $\sigma_\Phi = 2^\circ$.

Samples of Gaussian error with these standard deviations were added to body-fixed frame ($B_4$) elevation and azimuth measurements calculated as described in Section \ref{sec:noiseless}. These were converted to DOA measurements referenced to the INS frame $B_3$. The SDP+O algorithm was used to obtain $\bm{\overline{R}}$ and $\bm{\bm{\overline{t}}}$ using the agents' position coordinates in their respective navigation frames and the noisy DOA values. The reconstructed trajectory $\bm{\overline{p_B^{A_1}}}$ is plotted in {\color{black}Figure \ref{fig:frame_with_legend}} with the dotted black line. Position data of the reconstructed trajectory $\bm{\overline{p_B^{A_1}}}$ are tabulated {\color{black} in  \ref{tab:data}.}

\begin{remark}

The accuracy of the SDP+O solution in the noiseless case was observed to deteriorate when the condition number of the true solution for $\bm{X}$ is high. This is due to {\color{black}a form of inherent regularisation} in the SDP solver Yalmip \cite{Lofberg2004}. When the approximate magnitude of the norm $||\bm{t_{A_1}^{B_2}}||$ is known, one {\color{black}approach} is to straightforwardly introduce a scaling coefficient before entries $t_i$ for $i=1, 2, 3$ in equations (\ref{eq:xzlinkb}) and (\ref{eq:yzlinkb}) equal to the approximate norm of $||\bm{t_{A_1}^{B_2}}||$. 

Furthermore, if an approximation of $\bm{t_{A_1}^{B_2}}$ is known a priori as $\bm{\widetilde{t}_{A_1}^{B_2}}$, the following controlled shifting algorithm may be applied to reduce the effect of regularisation error. We define shifted positions $\bm{{p_{s}}_B^{B_2}}(k) = \bm{p_B^{B_2}}(k) - \bm{\widetilde{t}_{A_1}^{B_2}}$. By substituting $\bm{{p_{s}}_B^{B_2}}$ for $\bm{p_B^{B_2}}$ in the {\color{black}SDP}, the vector $\bm{\overline{t}}$ obtained through SDP is an estimate of 
$\bm{t_{A_1}^{B_2}} - \bm{\widetilde{t}_{A_1}^{B_2}}$.
\end{remark}
\begin{figure}
\centering
\includegraphics[width=1\linewidth]{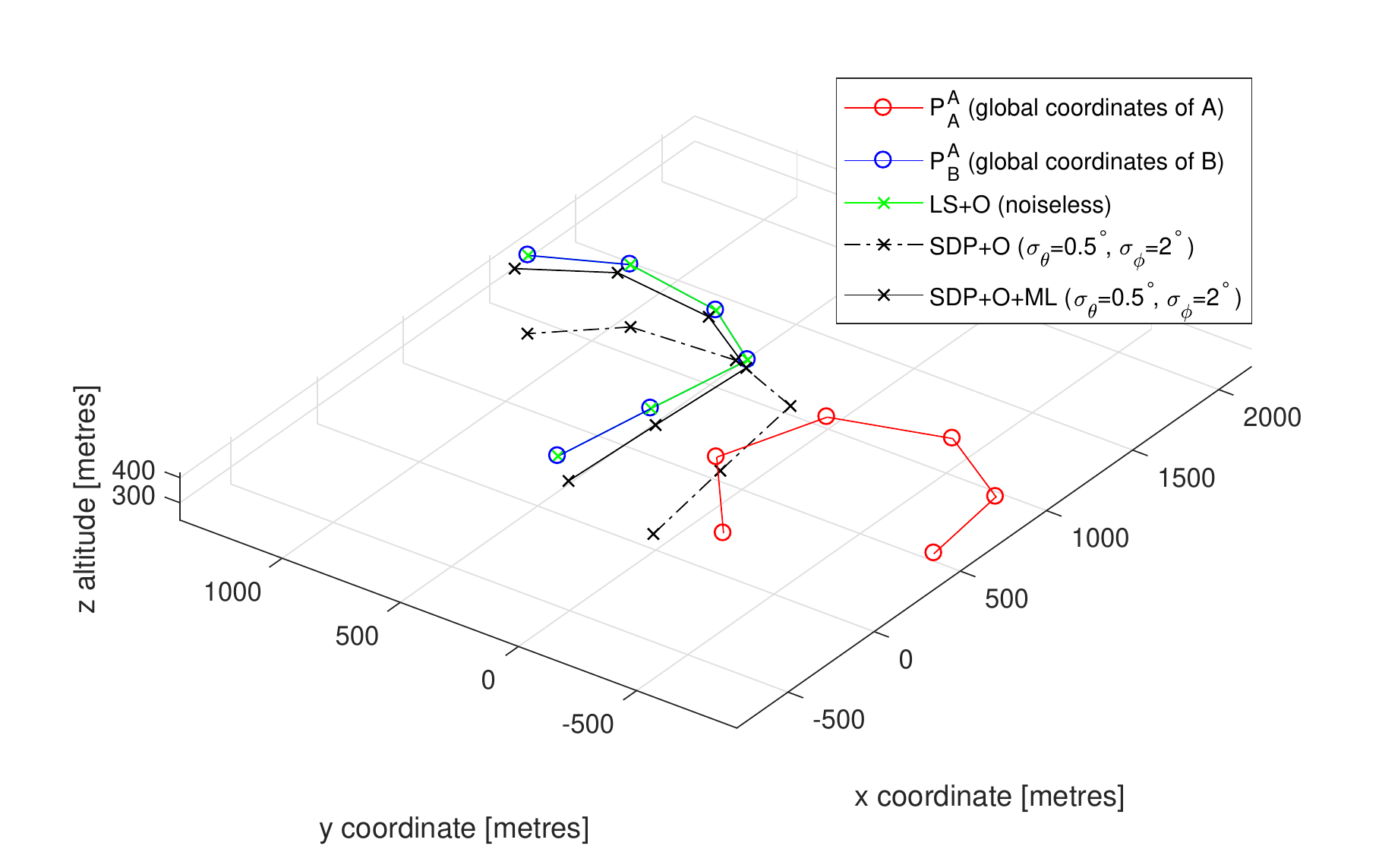}
\caption{\color{black}Recovery of global coordinates of Agent B for recorded trajectories. Errors are $\sigma_\theta = 0.5^\circ$ and $\sigma_\phi = 2^\circ$ with respect to body fixed frame for the DOA measurements}
\label{fig:frame_with_legend}
\end{figure}
\section{Maximum Likelihood Estimation}
\label{sec:mle}

This section presents a maximum likelihood estimation (ML) method to refine estimates $\bm{\overline{R}}$ and $\bm{\overline{t}}$ {\color{black}obtained using the SDP+O algorithm. The MLE refinement uses the series of DOA measurements expressed with respect to the body-fixed frame $B_4$, and the known values for $\sigma_\Theta$ and $\sigma_\Phi$ describing the expected distribution of DOA errors. A non-linear log-likelihood function for DOA measurement error is derived. The minimum of the log-likelihood function cannot be found analytically, so we employ an iterative gradient descent approach instead.

\subsection{Likelihood Function Derivation}
\label{ssec:likelihood_derivation}

In this section, DOA values are always expressed with respect to the body-fixed frame of Agent B ($B_4$) to exploit the independence of azimuth and elevation measurement errors. This is a change from Sections \ref{sec:noiseless} and \ref{sec:noisy_measurements}, in which DOA measurements were generally expressed with respect to the local INS frame $B_2$. {\color{black}The transformation between coordinate frames $B_2$ and $B_4$ is known to Agent B}.





Suppose body-fixed frame measurements of azimuth and elevation {\color{black}$\Theta(k)$ and $\Phi(k)$ are contaminated by zero mean Gaussian noise as follows:
\begin{itemize}
\item $\widetilde{\Theta}(k) = \Theta(k) + \xi_\Theta, \quad \xi_\Theta \sim N(0,{\sigma_\Theta}^2)$
\item $\widetilde{\Phi}(k) = \Phi(k) + \xi_\Phi, \quad \xi_\Phi \sim N(0,{\sigma_\Phi}^2)$
\end{itemize}}


To calculate noiseless azimuth and elevation measurements, an expression must be derived for the position of Agent A in Agent B's body-fixed frame $B_4$. Observe that
\begin{align}
\bm{\overline{p_A^{B4}}}(k) 
&= \bm{R_{B_2}^{B_4}}(k) (\bm{R_{A_1}^{B_2}} \bm{p_A^{A_1}}(k) + \bm{t_{A_1}^{B_2}}) + \bm{t_{B_2}^{B_4}}(k)
\end{align}
To help distinguish coordinate reconstructions based on estimates of $\bm{\bm{\overline{R}}}$ and $\bm{\overline{t}}$ from true coordinates, reconstructed positions will be explicitly expressed as functions of $\bm{\bm{\overline{R}}}$ and $\bm{\overline{t}}$:
\begin{align}
\bm{\overline{p_A^{B4}}}(k, \bm{\bm{\overline{R}}}, \bm{\overline{t}}) = \bm{R_{B_2}^{B_4}}(k) (\bm{\bm{\overline{R}}} \bm{p_A^{A_1}}(k) + \bm{\overline{t}}) + \bm{t_{B_2}^{B_4}}(k)
\end{align}
By definition of azimuth and elevation in Section \ref{sec:problem_def}:
\begin{align}
\theta_{B_4}(k, \bm{\bm{\overline{R}}}, \bm{\overline{t}}) = \, & \arcsin\Big(\frac{\bm{\overline{p_A^{B4}}}(k, \bm{\bm{\overline{R}}}, \bm{\overline{t}})_{z}}{||\bm{\overline{p_A^{B4}}}(k, \bm{\bm{\overline{R}}}, \bm{\overline{t}})||}\Big) \\
\phi_{B_4}(k, \bm{\bm{\overline{R}}}, \bm{\overline{t}}) = \, & \text{atan2}\Big(\bm{\overline{p_A^{B4}}}(k, \bm{\bm{\overline{R}}}, \bm{\overline{t}})_{y} \: , \: \bm{\overline{p_A^{B4}}}(k,\bm{\bm{\overline{R}}},\bm{\overline{t}})_{x}\Big)
\end{align}
where $\bm{\overline{p_A^{B4}}} = [\bm{\overline{p_A^{B4}}}_{x}, 
\bm{\overline{p_A^{B4}}}_{y}, 
\bm{\overline{p_A^{B4}}}_{z}]^\top$.
The likelihood function for the set of DOA measurements is defined as follows:
\begin{align}
&\bm{\mathcal{L}}(\bm{p_A^{A_1}},\bm{p_B^{B_2}}|\bm{\bm{\overline{R}}}, \bm{\overline{t}}) \nonumber \\
&= \frac{1}{\sigma_\Theta \sqrt{2\pi}} \prod_{k=1}^{K} \exp \Big[ - \frac{(\widetilde{\theta}_{B_4}(k)-\theta_{B_4}(k, \bm{\bm{\overline{R}}}, \bm{\overline{t}}))^2}{2\sigma_\Theta^2} \Big] \nonumber \\
&\times \frac{1}{\sigma_\Phi \sqrt{2\pi}}  \prod_{k=1}^{K} \exp \Big[ - \frac{(\widetilde{\phi}_{B_4}(k)-\phi_{B_4}(k, \bm{\bm{\overline{R}}}, \bm{\overline{t}}))^2}{2\sigma_\Phi^2}\Big]
\end{align}
{\color{black}It can be shown that maximising $\bm{\mathcal{L}}(\bm{p_A^{A_1}},\bm{p_B^{B_2}}|\bm{\bm{\overline{R}}}, \bm{\overline{t}})$ is equivalent to} minimising
\begin{equation}
\sum_{k=1}^{K} \Big[ \frac{(\widetilde{\theta}_{B_4}(k)-\theta_{B_4}(k, \bm{\bm{\overline{R}}}, \bm{\overline{t}}))^2}{2\sigma_\Theta^2} +\frac{(\widetilde{\phi}_{B_4}(k)-\phi_{B_4}(k, \bm{\bm{\overline{R}}}, \bm{\overline{t}}))^2}{2\sigma_\Phi^2} \Big]
\label{eq:ml}
\end{equation}


\subsection{Gradient descent and adaptive step size}

Possible parametrisations for the rotation matrix $\bm{\overline{R}}$ include Euler angles, quaternion representation and Rodrigues rotation formula. In this paper we parametrise $\bm{\overline{R}}$ by a 3-vector of Euler angles, and $\bm{\overline{t}}$ is a 3-vector. This defines a mapping from $\mathbb{R}^6 \rightarrow \bm{\overline{R}}, \bm{\overline{t}}$, and the gradient of (\ref{eq:ml}) can be expressed as a vector in $\mathbb{R}^6$. The log-likelihood function is non-linear with respect to this $\mathbb{R}^6$ parametrisation of $\bm{\bm{\bm{\overline{R}}}}$ and $\bm{\bm{\overline{t}}}$. As a result, this function may be non-convex, meaning the equation $\mathcal{D} \log \mathcal{L} = 0$ may have multiple solutions, with only one of these being the global minimum. A gradient descent algorithm is therefore initialised using the result of the SDP+O method, and is used to converge towards a local minimum, which it is hoped will be the global minimum or close to it.

Instead of selecting a constant step size, which may lead to overshooting local minima or excessive computation time, an adaptive step size approach is adopted. {\color{black} The backtracking line search algorithm discussed in \cite{stanimirovic2010accelerated} selects an optimally large step size satisfying a constraint placed on the average gradient over the step.}

\subsection{Example of ML refinement of SDP+O solution}

In this subsection, we demonstrate the benefits of maximum likelihood estimation. ML was performed using the real flight trajectory data presented in Sections \ref{sec:noiseless} and \ref{sec:noisy_measurements}. The resulting reconstructed trajectory $\overline{\bm{p_A^{B_2}}}$ is presented in Figure \ref{fig:frame_with_legend} as the solid black line, and its coordinates are {\color{black} tabulated in Table \ref{tab:data}}. Additionally, in this section we present the decrease {\color{black}and convergence} in the value of the negative log-likelihood function (\ref{eq:ml}), frame rotation error and reconstructed position error\footnote{Metrics are defined in the sequel, see Section \ref{ssec:error_metrics} below.} over successive iterations of the gradient descent algorithm in Figures \ref{fig:mle_funct_val}, \ref{fig:mle_rotation} and \ref{fig:mle_position}}. 

{\color{black}The error in INS frame rotation is reduced by over 60\%, and the reconstructed position error of Agent B is reduced by over 70\% by iterating the gradient descent algorithm. This represents a significant gain with respect to the SDP+O estimate, which served as the initialisation point of the gradient descent. Monte Carlo simulations covering a large set of trajectories are presented in Section \ref{sec:simulations}}.


 \begin{figure}[t]
 \centering
 \includegraphics[width=\linewidth]{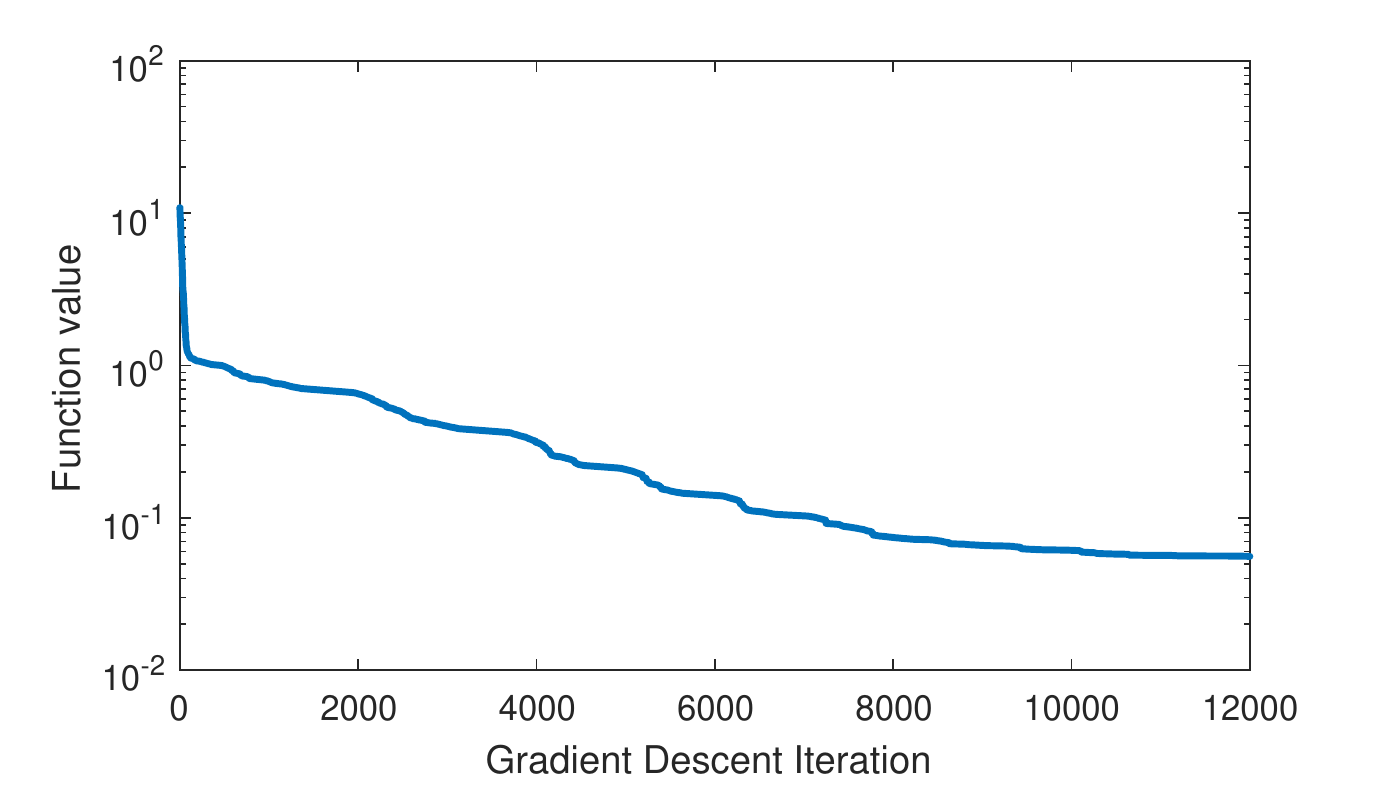}
 \caption{Convergence of negative log-likelihood function using ML for real trajectory pair}
 \label{fig:mle_funct_val}
 \end{figure}

\begin{figure}[t]
\centering
\includegraphics[width=\linewidth]{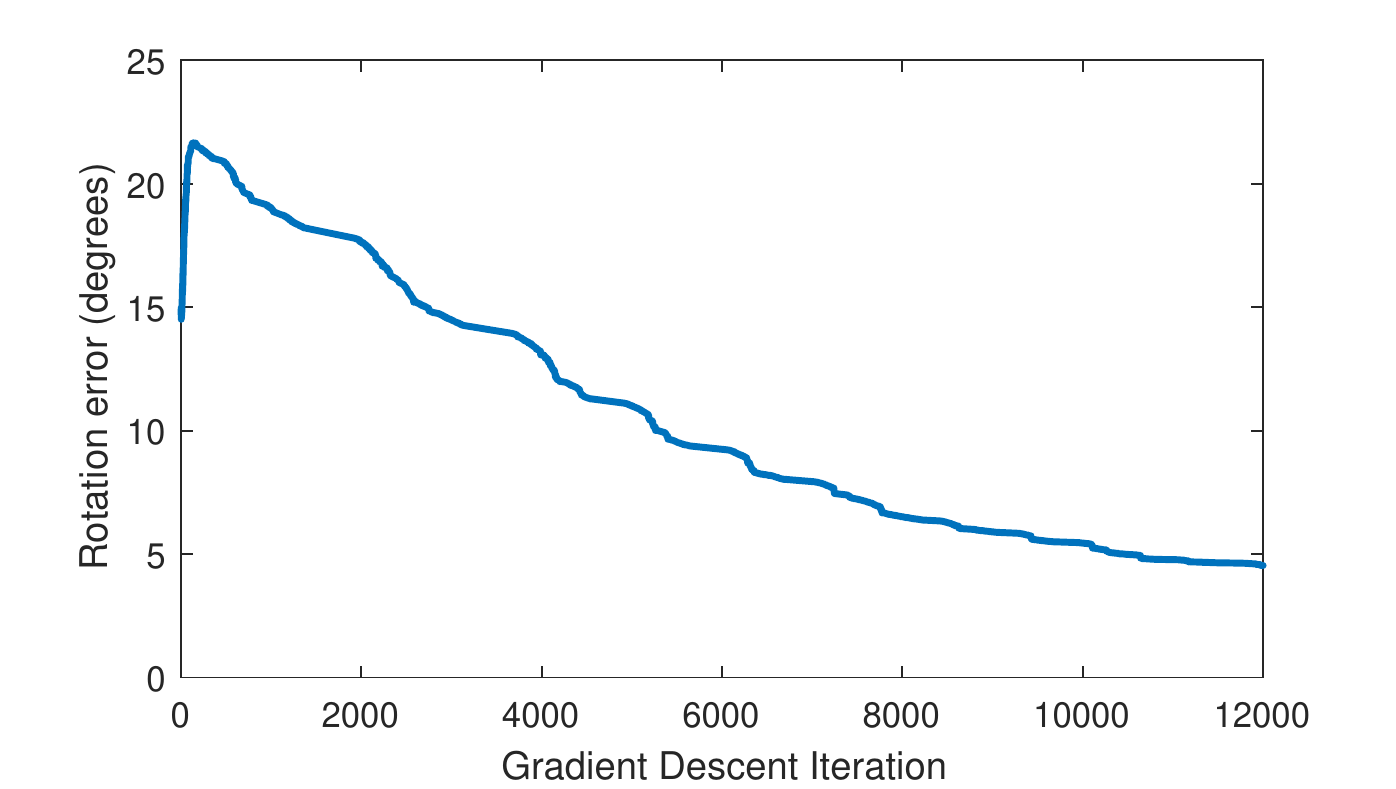}
\caption{Improvement in rotation error in degrees using ML for real trajectory pair}
\label{fig:mle_rotation}
\end{figure}
\begin{figure}[t]
\centering
\includegraphics[width=\linewidth]{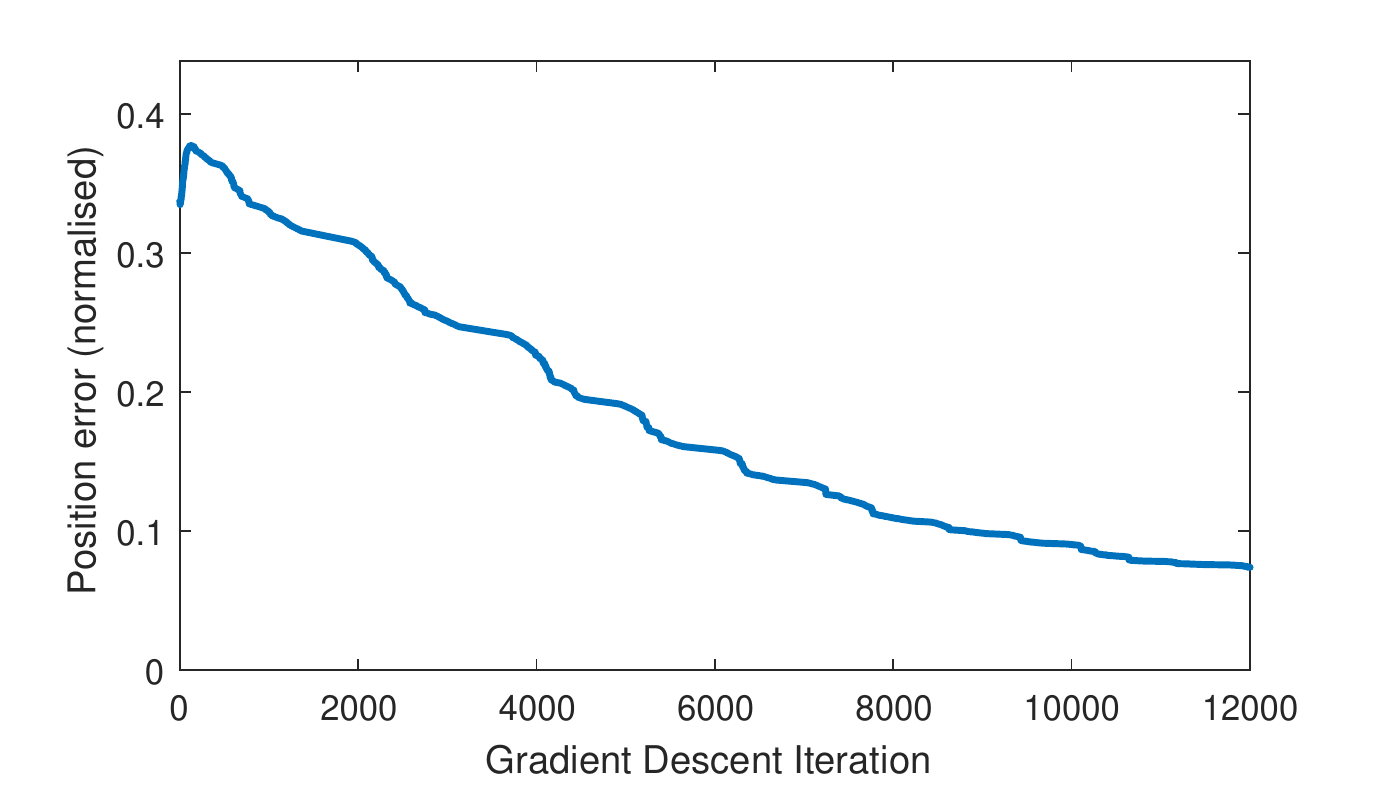}
\caption{Improvement in reconstructed position error using ML for real trajectory pair}
\label{fig:mle_position}
\end{figure}



\color{black}
\section{Simulation Results}\label{sec:simulations}

In this section, metrics are defined for performance evaluation of the localisation algorithm. Monte Carlo simulations of realistic\footnote{These trajectories satisfy a set of assumptions detailed in Section \ref{ssec:rand_traj}.} flight trajectories are performed to evaluate the effect of errors in body-fixed frame azimuth and elevation measurements. {\color{black}We also investigate the expected incremental improvement in localisation accuracy as the number of DOA measurements $K$ increases.} Finally, trajectories of Agents A and B which are unsuitable for localisation of Agent B are discussed.

{\color{black}In the preliminary conference paper \citep{RUSSELL20178019}, we compared the performance of the LS+O and SDP+O methods 
The LS+O method collapsed when small amounts of noise were introduced to DOA measurements, whereas rotation error increased linearly with respect to DOA measurement noise when using the SDP+O method. 
The SDP+O method is the superior method, and there is no reason to employ LS+O.}

\subsection{Metrics for error in $\bm{\overline{R}}$ and $\bm{\overline{t}}$}\label{ssec:error_metrics}
This paper uses the geodesic metric for rotation \citep{Huynh2009}.
All sequences of rotations in three dimensions can be expressed as one rotation about a single axis \citep{palais2009disorienting}.
The geodesic metric on $SO(3)$ defined by
\begin{equation}
d(\bm{R_1,R_2}) = \arccos\bigg(\frac{\text{tr}(\bm{R_1}^\top \bm{R_2})-1}{2}\bigg)
\end{equation}
is the magnitude of angle of rotation about this axis \citep{7515271}. Where $\bm{R_A^B}$ is known, the error of rotation $\bm{\overline{R}}$ is defined as $d(\bm{\overline{R}},\bm{R_A^B})$. 
Position error is defined as the average Euclidian distance between true global coordinates of Agent B, and estimated global coordinates over the $K$ measurements taken, divided (to secure normalisation) by the average distance between aircraft.
\begin{equation}
\bm{\overline{p_B^{A_1}}}(k) =  \bm{\overline{R}}^\top  (\bm{p_B^{B_2}} - \bm{\overline{t}})
\end{equation}
\begin{equation}
error(\bm{\overline{p_B^{A_1}}}) = \frac{\sum_k ||\bm{\overline{p_B^{A_1}}}(k) - \bm{p_B^{A_1}}(k)||}{Kd}
\end{equation}
where $d$ represents the average distance between aircraft.

\subsection{Monte Carlo Simulations on Random Trajectories using SDP+O and ML}\label{ssec:rand_traj}

\begin{figure}
\includegraphics[width = \linewidth] {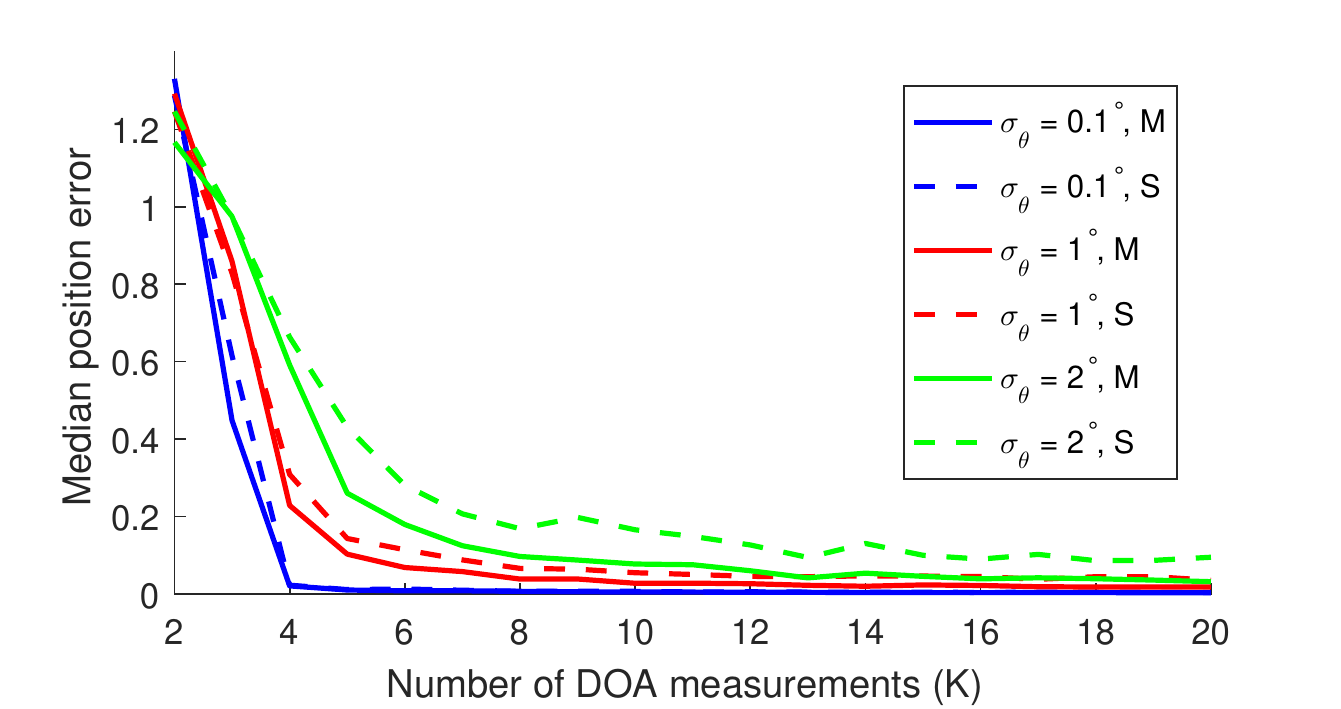}
\caption{Median $error(\overline{\bm{p_B^{A_1}}})$ vs. number of DOA measurements used to solve SDP+O (S) and SDP+O+ML (M) from $K = 2$ to $K = 20$.}

\label{fig:position}
\end{figure}

\begin{figure}
\includegraphics[width = \linewidth] {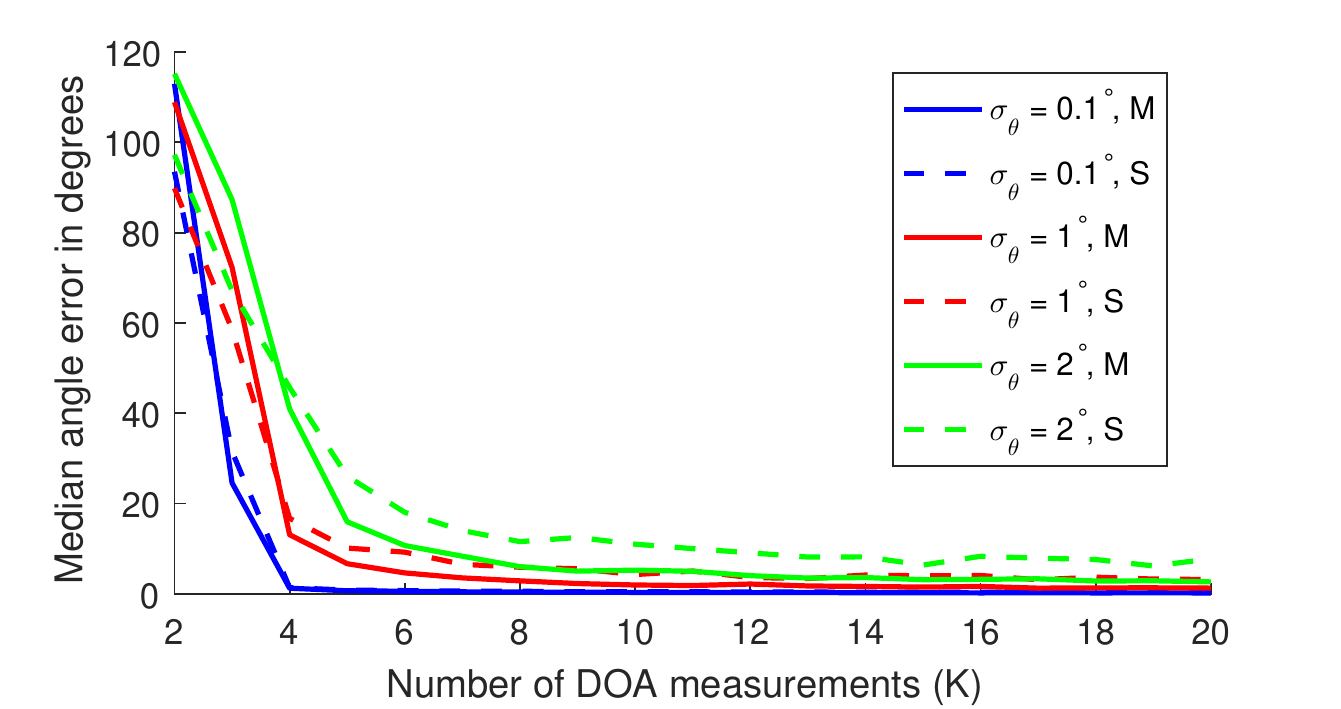}
\caption{Median $d(\bm{\overline{R}},{\bm{R^B_A}})$ vs. number of DOA measurements used to solve SDP+O (S) and SDP+O+ML (M) from $K = 2$ to $K = 20$}
\label{fig:angles}
\end{figure}

In this subsection, we summarise the results of Monte Carlo simulations to evaluate the expected performance of the SDP+O method and the SDP+O+ML method.

Pairs of realistic trajectories for Agents A and B are generated in accordance with the following assumptions:
\begin{itemize}
	\item Initial horizontal separation of 800m
	\item Initial vertical separation of 50m (initial altitudes of 300 and 350 metres)
	\item Average speed of 50 metres per second (97 knots)
	\item Measurements taken every 5 seconds
	\item Initial compass heading of each UAV in the $xy$ plane follows a uniform distribution from 0 to 360 degrees. This is expressed as $h \sim U(0, 360)$ degrees
	\item Every 5 seconds, the recorded direction of each UAV in the $xy$ plane changes by $\delta$ degrees where $\delta \sim N(c, 30^2)$. The value $c \sim U(-40, 40)$ is constant for a given trajectory and corresponds to an average curve for a trajectory
	\item At each time measured, the rate of climb of each UAV in degrees is $RoC \sim N(0, 5^2)$
\end{itemize}

{\color{black} To represent the drift in the INS of Agent B}, rotations $\bm{R_{A_1}^{B_2}}$ were generated by independently sampling three Euler angles $\alpha, \beta, \gamma$ where $\alpha, \beta, \gamma \sim U(-\pi, \pi)$, and translations $\bm{t_{A_1}^{B_2}} = [t_1, t_2, t_3]^\top$ were generated by sampling entries $t_1, t_2, t_3 \sim U(-600, 600)$.

As discussed in Section \ref{ssec:example_noisy}, we assume the standard deviation in elevation measurement error in the body-fixed frame $B_4$ is four times the standard deviation in azimuth measurement error in $B_4$, i.e. $\sigma_\Phi = 4 \sigma_\Theta$. We vary the DOA error by $\sigma_\Theta = [0.1^\circ \, , \, 1^\circ \, , \, 2^\circ]$. Errors in the order of $\sigma_\Theta = 0.1^\circ$ are representative of an optical sensor, whereas the larger errors are representative of antenna-based (RF) measurements. 

For each value of $\sigma_\Theta$ studied, and for each number of DOA measurements $K$ from 2 to 20, we simulated 100 unique and realistic UAV trajectory pairs (Agent A and Agent B). For each trajectory pair, localisation was performed using the SDP+O method, and metrics $d(\bm{\overline{R}},{\bm{R^B_A}})$ and $error(\bm{\overline{P_B^{A_1}}})$, were calculated. The ML method was then used to enhance the result of the SDP+O method, and the error metrics were recalculated. After all simulations were completed, the median\footnote{\color{black}For asymmetric distributions such as errors (which are nonnegative by definition and contain extreme outliers), the median is a superior measure of central tendency than the mean \cite{boslaugh2012statistics,stat_ref}.} values of $d(\bm{\overline{R}},{\bm{R^B_A}})$ and $error(\bm{\overline{P_B^{A_1}}})$ for both the SDP+O and SDP+O+ML methods were calculated across each set 100 simulations. The results of the Monte Carlo simulations are plotted in Figures \ref{fig:position} and \ref{fig:angles}. 

{\color{black} Median $d(\bm{\overline{R}},{\bm{R^B_A}})$ and $error(\bm{\overline{P_B^{A_1}}})$ errors decrease significantly when 4 or more DOA measurements ($K$) are used. Both metrics show an asymptotic limit to performance across all three noise levels as the number of DOA measurements ($K$) increases. Median rotation error $d(\bm{\overline{R}},{\bm{R^B_A}})$ and $error(\overline{\bm{p_B^{A_1}}})$ appear to exhibit similar {\color{black}asymptotic performance gain} over the number of DOA measurements $K$ up to 20.}

\begin{remark}
{\color{black} The maximal parametrisation for $\bm{\overline{R}}$, $\bm{\overline{t}}$ consists of 12 entries, and the largest set of independent quadratic constraints consists of 6 relations. Polynomial equation sets of $n$ independent relations in $n$ unknowns will generically have multiple solutions if some relations are quadratic. The addition of a scalar measurement generically yields a unique solution. 
Therefore 4 DOA measurements are required to obtain the minimum of $(12-6)+1=7$ scalar measurements.}
\end{remark}

\subsection{Unsuitable trajectories for localisation}
\label{ssec:nongeneric_traj}

In this subsection we are motivated to identify trajectories of Agents A and B which may lead to multiple solutions for $\bm{\overline{R}}$ and $\bm{\overline{t}}$ in the noiseless case, and consequently unreliable solutions in the noisy case. 
One can prove that if Agent A's motion is restricted to a plane, a set of three columns of matrix $\bm{A}$ in Eqn. (\ref{eq:Axb}) become linearly dependent, and therefore a unique solution cannot be obtained using the LS method.

When quadratic constraints are included in the SDP+O method,
rank deficiency of $\bm{A}$ no longer automatically implies the existence of multiple solutions. For example, the SDP+O method obtains a unique solution if Agent A's motion is planar and Agent B's motion is arbitrary. Multiple solutions nevertheless can exist for certain non-generic unsuitable trajectories for the SDP+O method.

\begin{figure}
\centering
\includegraphics[width=\linewidth]{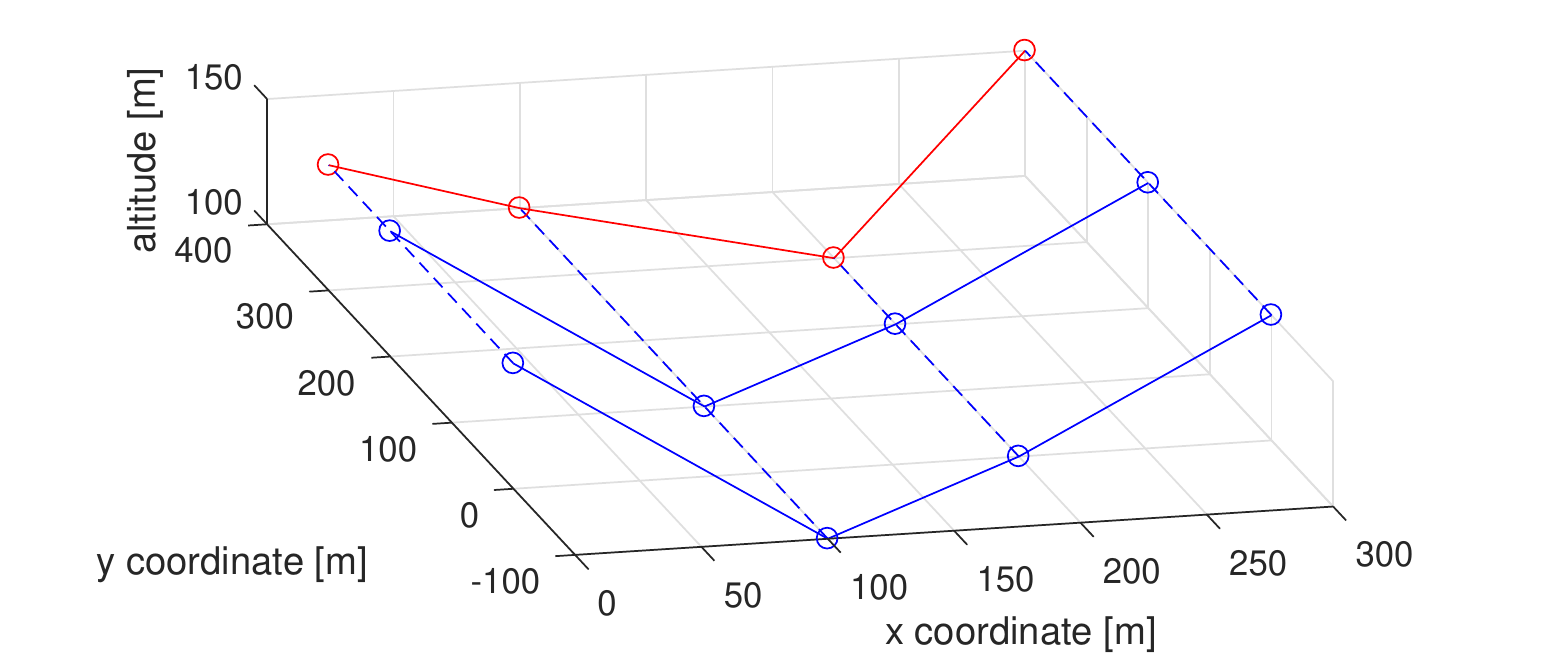}
\caption{Illustration of trajectory pairs leading to equal DOA measurements (disconnected blue lines) over $K$ measurement instants. SDP+O+ML algorithm unable to discern distance from which Agent B (solid blue) observes Agent A (red).}
\label{fig:parallel_measurements}
\end{figure}

{\color{black}We begin by considering the case where DOA measurements expressed with respect to the Local INS frame $B_2$ are equal at each time instant. 
This is illustrated by an example in Figure \ref{fig:parallel_measurements}.
{\color{black}A similar problem is expected in the far field case, where the distance between Agents A and B is sufficiently large that DOA measurements become approximately equal despite each Agent's trajectory remaining arbitrary.}
In these cases, 
multiple solutions exist for 
$\bm{t_{A_1}^{B_2}}$.}




\begin{figure}
\centering
\includegraphics[width=\linewidth]{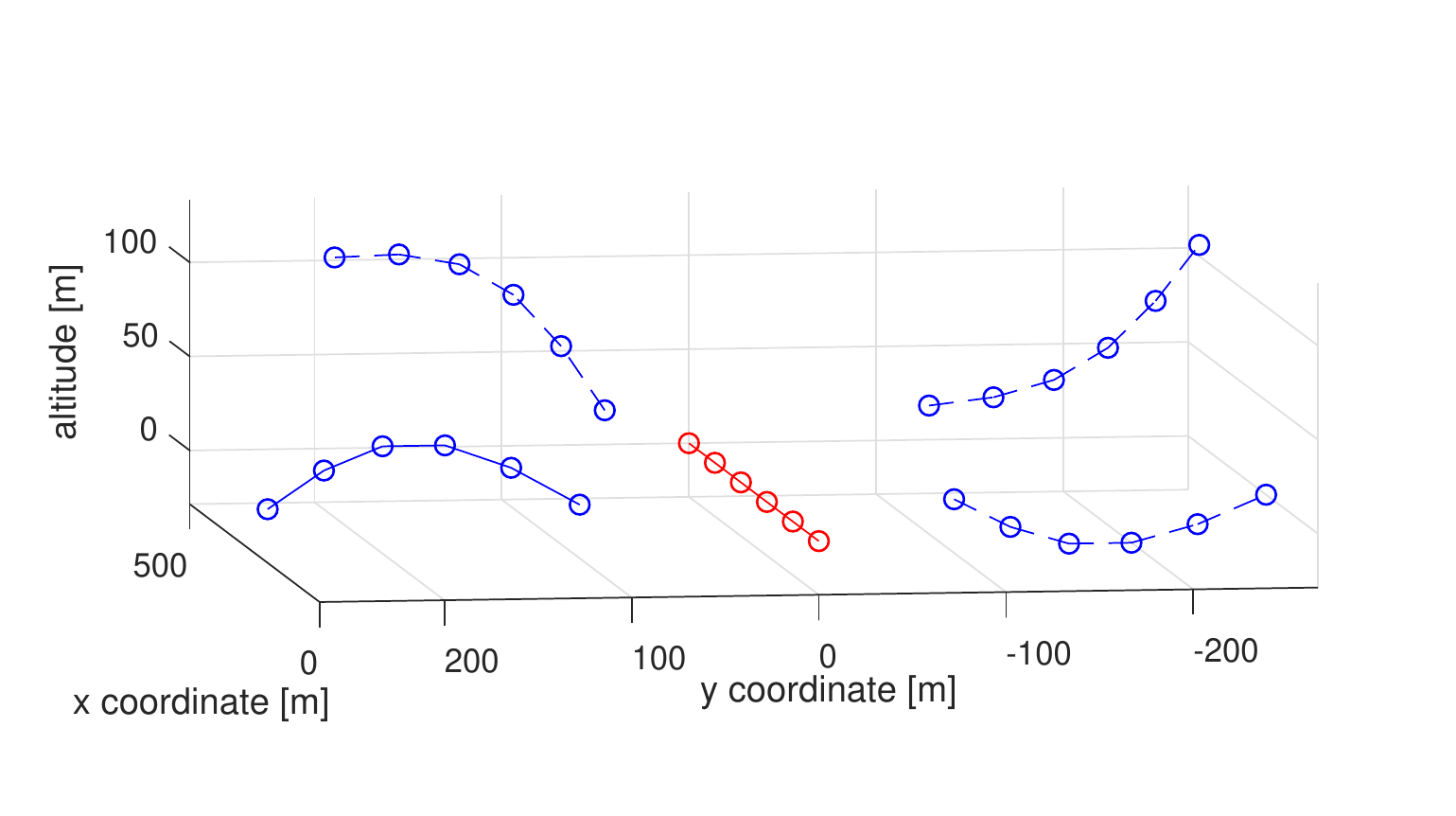}
\caption{{\color{black}Illustration of straight line motion of Agent A (red, trajectory given by  $(x(t),y(t),z(t)) = (100t, 0, 0)$). Agent B (blue) observes Agent A through an unaligned INS frame. 
In this figure, the solid blue trajectory is the actual path of Agent B. However, each dotted blue line is also an admissible solution.}}
\label{fig:agent_a_linear}
\end{figure}

Multiple solutions may also arise if Agent A's trajectory appears similar from multiple perspectives. The localisation process may be incapable of determining the direction from which DOA measurements were taken with respect to the global frame. For example, if Agent A follows a straight line, a set of recorded DOA measurements may be achieved by viewing Agent A from any direction in a circle perpendicular to Agent A's motion and centred at Agent A's trajectory. This is illustrated in Figure \ref{fig:agent_a_linear}.

\section{Three-Agent Extension and Beyond}

This section explores {\color{black}in a preliminary way} how the SDP+O+ML algorithm may be extended to localise two GPS-denied agents efficiently. Trivially, each GPS-denied aircraft could measure DOA of the GPS-equipped agent's broadcast of its position, and use the SDP+O+ML algorithm independently of each other to estimate drift in their local frames. We are motivated to determine whether a \textit{trilateral}\footnote{\color{black}In this section we relax the condition preventing GPS-denied agents from broadcasting signals} algorithm may be more resilient to DOA measurement error and/or unsuitable trajectories, and may perhaps require fewer DOA measurements from each aircraft than simply repeating the two-agent localisation algorithm with each GPS-denied agent. We introduce a GPS-denied Agent C, whose local INS frame has rotation and translation parameters $\bm{R_{A_1}^{C_2}}$ and $\bm{t_{A_1}^{C_2}}$ with respect to the global frame. We conclude this section by discussing the challenges involved in generalising our findings to arbitrary $n$-agent networks.

\subsection{Measurement process in three-agent network}
To describe measurements within a network of more than two agents, one minor notation change is required:
\begin{itemize}
\item DOA measurements made by Agent I towards Agent J will henceforth be expressed in the INS coordinate frame of Agent I as ($\theta_{I_2}^J$, $\phi_{I_2}^J$)
\end{itemize}
At each time instant $k$ in the discrete-time process: 
\begin{itemize}
\item Agents A and B interact as per the two-agent case.
\item Agent C receives the broadcast of Agent A's global coordinates, and measures this signal's DOA with respect to frame $C_2$, which we denote ($\theta_{C_2}^A$, $\phi_{C_2}^A$). 
\item Agent C broadcasts its position with respect to its INS frame $\bm{p_C^{C_2}}$, as well as the measurement ($\theta_{C_2}^A$, $\phi_{C_2}^A$) to 
Agent B, who also takes a DOA measurement towards Agent C. This measurement is denoted ($\theta_{B_2}^C$, $\phi_{B_2}^C$). 
\end{itemize}

All DOA and position measurements are therefore relayed to Agent B, who performs the localisation algorithm presented in this section.

\subsection{Forming system of linear equations in three-agent network}

In 
Section \ref{sec:noiseless}, the linear system $\bm{A\Psi} = \bm{b}$ was formed using relations stemming from the collinearity of the vector ($\bm{p_A^{B_2}} - \bm{p_B^{B_2}}$), and the vector in the direction of DOA measurement ($\theta_{B_2}^A$, $\phi_{B_2}^A$). We refer to this system of equations as $\bm{S_{AB}}$, where the subscript references the agents involved. A similar system $\bm{S_{AC}}$ can be constructed independently using Agent C's DOA measurements towards Agent A and $\bm{p_C^{C_2}}$.

In the three-agent network, Agent B also measures the DOA towards Agent C's broadcast, with respect to Agent B's local INS frame $B_2$. To exploit the collinearity of the vectorial representation of the DOA measurement ($\theta_{B_2}^C$, $\phi_{B_2}^C$) and ($\bm{p_C^{B_2}} - \bm{p_B^{B_2}}$), an expression for the position coordinate vector $\bm{p_C^{B_2}}$ is required. As achieved in equations (7) and (8) in Section \ref{sec:noiseless}, this position may be expressed in terms of entries of $\bm{R_{C_2}^{B_2}}$ and $\bm{t_{C_2}^{B_2}}$, and the linear system $\bm{S_{BC}}$ may be defined similarly to $\bm{S_{AB}}$ in Section \ref{sec:noiseless}. 
Systems $\bm{S_{AB}}$, $\bm{S_{AC}}$ and $\bm{S_{BC}}$ can be assembled, forming a large system of linear equations $\bm{S_{ABC}}$ with 36 scalar unknowns (9 rotation matrix entries and 3 translation vector entries for each distinct agent pair).

{\color{black}At each time instant $k$ for $k = 1, ..., K$, two linear equations are obtained from each DOA measurement of
($\theta_{B_2}^A$, $\phi_{B_2}^A$), ($\theta_{C_2}^A$, $\phi_{C_2}^A$) and ($\theta_{B_2}^{C_2}$, $\phi_{B_2}^{C_2}$). As a result, 6 linear equations are obtained at each time instant. Performing the measurement process 6 times $(K = 6)$ produces 36 linear equations. Generically, in the noiseless case, a unique solution therefore exists for $K=6$ time instants. When using only the LS method, 
the three-agent localisation problem requires the same minimum number of time instants as solving two independent two-agent localisation problems concurrently, yet requires more DOA measurements than the sum of the number of measurements required in two separate two-agent localisation problems. However, quadratic relationships between $\bm{R_{A_1}^{B_2}}$, $\bm{t_{A_1}^{B_2}}$, $\bm{R_{A_1}^{C_2}}$, $\bm{t_{A_1}^{C_2}}$, $\bm{R_{B_2}^{C_2}}$ and $\bm{t_{B_2}^{C_2}}$ significantly reduce the required number of time instants $(K)$ at which measurements occur.}

\subsection{Quadratic constraints in three-agent network and example} \label{ssec:3_agent_constraints}

 In the two-agent case, 21 linearly independent quadratic constraints were identified in order to determine the rotation and translation between two frames. {\color{black}In the three-agent case, when the underlying undirected graph is a clique}, we identify $21 \times 3 = 63$ linearly independent quadratic constraints using rotation matrix properties of $\bm{R_{A_1}^{B_2}}$, $\bm{R_{A_1}^{C_2}}$, and $\bm{R_{B_2}^{C_2}}$.

 In the three-agent case, additional quadratic constraints can be identified due to relationships between rotated and translated reference frames.
 \begin{itemize}
 \item Applying the rotation $\bm{R_{A_1}^{C_2}}$ is equivalent to applying rotations $\bm{R_{A_1}^{B_2}}$ and $\bm{R_{B_2}^{C_2}}$ successively. This relationship is expressed in equation (\ref{eq:constr_eq_1}). Setting the entries of the left hand side to zero yields 9 quadratic constraints.
 \item The difference between the global frame representation of vectors $\bm{t_{A_1}^{C_2}}$ and $\bm{t_{A_1}^{B_2}}$ is equal to the global frame representation of $\bm{t_{B_2}^{C_2}}$. This relationship is expressed in equation (\ref{eq:constr_eq_2}). Setting entries of the left hand side to zero yields 3 quadratic constraints.
 \end{itemize}
 \begin{align}
 \label{eq:constr_eq_1}
 \bm{R_{A_1}^{C_2}} - \bm{R_{B_2}^{C_2}} \bm{R_{A_1}^{B_2}} = \bm{0} \\
 \label{eq:constr_eq_2}
 (\bm{R_{C_2}^{A_1}}\bm{t_{A_C}^{C_2}} - \bm{R_{B_2}^{A_1}}\bm{t_{A_1}^{B_2}}) - \bm{R_{C_2}^{A_1}}\bm{t_{B_2}^{C_2}} = \bm{0}
 \end{align}

 As mentioned in Section IV-C, dependent constraints which are not linearly dependent are included to improve the accuracy of the SDP solver. The relationship between rotations described in equation (\ref{eq:constr_eq_1}) may be expressed in 3 distinct coordinate frames ($A_1$, $B_2$ and $C_2$), and hence $9 \times 3 = 27$ linearly independent quadratic constraints may be derived using the relationship in equation (\ref{eq:constr_eq_1}). Similarly, the relationship between translations described in equation (\ref{eq:constr_eq_2}) may be expressed in 3 distinct coordinate frames ($A_1$, $B_2$ and $C_2$), and hence $3 \times 3 = 9$ linearly independent quadratic constraints may be derived using the relationship in equation (\ref{eq:constr_eq_2}).
 In total, we have derived $(21 \times 3) + (9 \times 3) + (3 \times 3) = 99$ linearly independent quadratic constraints for a system of 36 unknown variables. These constraints may be expressed in inner-product form as performed in Section \ref{sec:noisy_measurements}.

Rank-relaxed semidefinite programming can be used to obtain solutions for each INS frame's rotation and translation with respect to the global frame, and the Orthogonal Procrustes algorithm can be applied to each individual resulting rotation matrix. This defines the three-agent SDP+O method.

\begin{figure}
\centering
\includegraphics[width=0.8\linewidth]{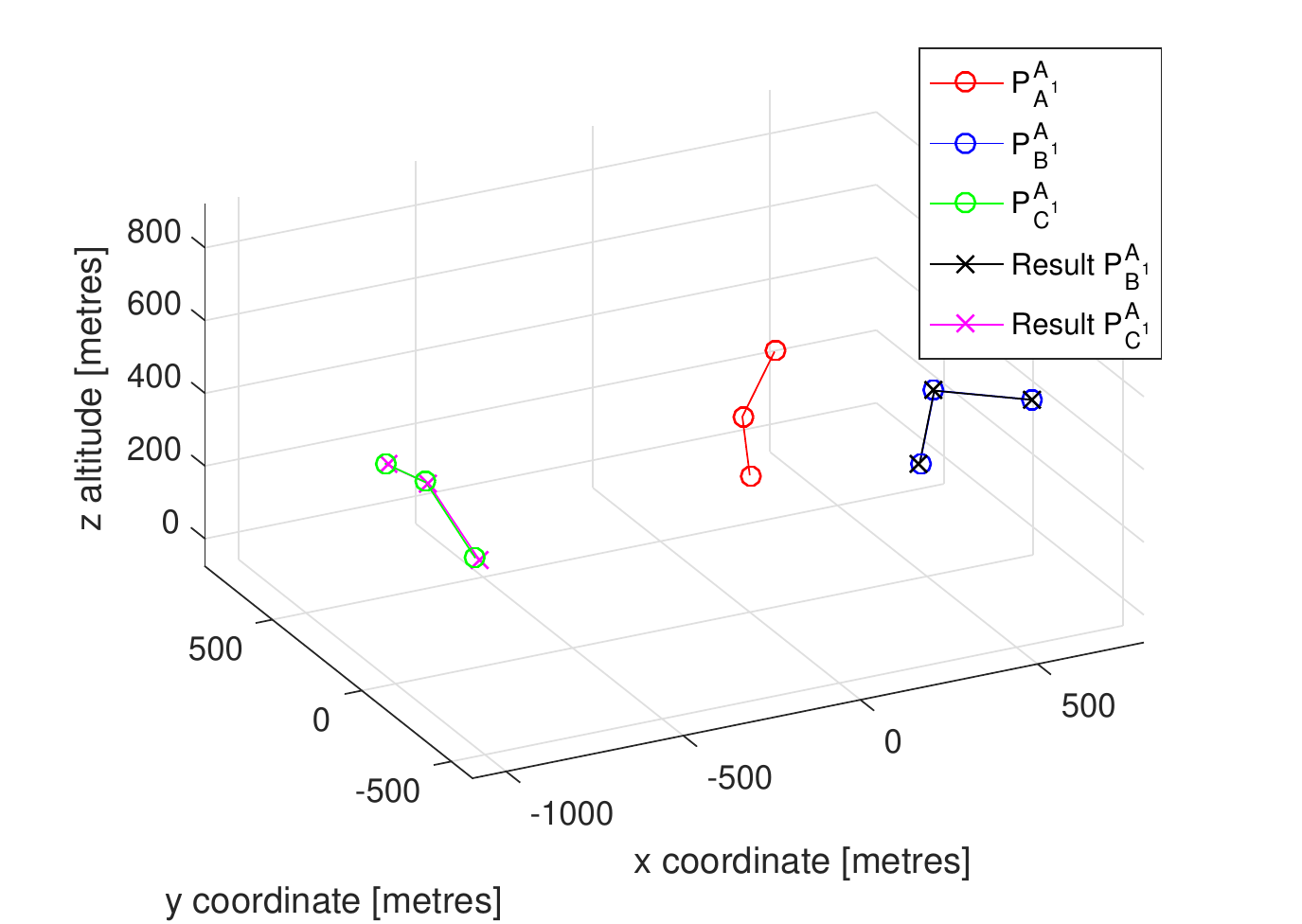}
\caption{Illustration of example of successful localisation within three-agent network in noiseless case for K = 3}
\label{fig:3_agents}
\end{figure}

To illustrate successful localisation in the three-agent case, realistic trajectories were defined for Agents A, B and C for $K = 3$ time instants. These are presented in Figure \ref{fig:3_agents}. Only Agents B and C were assigned random INS frame rotations and translations as prescribed in Section \ref{sec:simulations}, and the three-agent SDP+O method was used to obtain estimates of $\bm{R_{A_1}^{B_2}}$, $\bm{t_{A_1}^{B_2}}$, $\bm{R_{A_1}^{C_2}}$ and $\bm{t_{A_1}^{C_2}}$. {\color{black} Each directional measurement consists of two scalar measurements, and hence a total of $3\times 2 \times K = 18$ scalar measurements were obtained.} Localisation was successful, which demonstrates that only 3 time instants ($K=3$) are required for the three-agent SDP+O algorithm to obtain the exact solution in the noiseless case. Earlier, it was established that a minimum of 6 time instants were required to achieve a unique solution in the three-agent case using LS+O, and a minimum of 4 time instants were required to achieve a unique solution in the two-agent case using SDP+O. {\color{black}We have therefore demonstrated that a trilateral algorithm can achieve localisation of two GPS-denied agents in less measurement time instants than applying the bilateral algorithm twice independently. We note that this extension to three-agents is not applicable if the measurement graph is a tree because measurements are required between each pair of agents within the three-agent network.}

\subsection{Challenges in extension to $n$-agent networks}

{\color{black}While the results in Section VII-C demonstrate that localisation of two agents may be achieved in fewer measurement time instants $K$ when a three-agent extension is used, formalising an extension to arbitrary $n$-agent networks presents a significant theoretical challenge.

There exist comprehensive works such as \citep{7163542} and \citep{Zelazo2014RigidityTI} on bearing rigidity of an arbitrary network in $\mathbb{R}^2$ where all agents share the same reference frame, and some concepts are generalised to $\mathbb{R}^n$ in \citep{ZHAO2016334}. However, bearing rigidity theory for networks in $\mathbb{R}^n$ when agents do not share a reference frame is comparatively underdeveloped when compared to distance-based rigidity. For example, a general theory does not exist for the minimum number of measurements required for rigidity. 

We also note the risk of an explosion in computational complexity when extending our algorithm to $n$-agent networks. The relative pose of INS frames of any two agents linked by an edge in the underlying undirected measurement graph must be determined in order to use Eqn. (8). We cannot substitute entries of $\bm\Psi$ relating to the relative pose of two INS frames (such as entries of $\bm{R_{B_2}^{C_2}}$ and $\bm{t_{B_2}^{C_2}}$ in the three-agent case) with associated quadratic expressions using relationships such as Eqn. (38) or Eqn. (39), or else the objective in Eqn. (15) will cease to be quadratic and the SDP method will not be applicable. As a result, extending the algorithm to a large number of agents risks an exponential increase in the number of variables to be determined. It is not clear exactly how many measurements between the $n$ agents would be required to obtain a result in the noiseless case.}








\section{Conclusion}\label{sec:conclusion}
This paper studied a cooperative localisation problem between a GPS-denied and a GPS-enabled UAV. A localisation algorithm was developed in two stages. We showed that a linear system of equations built from six or more measurements yielded the localisation solution for generic trajectories. The second stage considered the inclusion of quadratic constraints due to rotation matrix constraints. Rank relaxed semidefinite programming was used, and the solution adjusted using the Orthogonal Procrustes algorithm. This gave the algorithm greater resilience to noisy measurements and unsuitable trajectories. Maximum likelihood estimation was then used to improve the algorithm's results. Simulations were presented to illustrate the algorithm's performance. Finally, an approach was outlined to extend the two-agent solution to a three agent network in which only one agent has global localisation capacity. {\color{black}Future work may include implementation on aircraft to perform localisation in real time and validate our Monte Carlo analysis on measurement noise. We also hope to extend our trilateral algorithm to larger networks by establishing further theory on bearing rigidity when agents do not share a common reference frame.}


\bibliographystyle{IEEEtran}
\bibliography{Literature2}
\section*{Appendix}

\subsection{Forms of $\bm{A}$ and $\bm{b}$}

{\color{black}The matrix $\bm{A}$ is defined in Section III-A as follows:
	\begin{align*}
	&\bm{A}(2k-1,1) = u_A(k) \sin(\phi(k)) \\
	&\bm{A}(2k-1,2) = v_A(k) \sin(\phi(k)) \\
	&\bm{A}(2k-1,3) = w_A(k) \sin(\phi(k)) \\
	&\bm{A}(2k-1,4) = 0 \\
	&\bm{A}(2k-1,5) = 0 \\
	&\bm{A}(2k-1,6) = 0 \\
	&\bm{A}(2k-1,7) = -u_A(k)\cos(\theta(k))\cos(\phi(k)) \\
	&\bm{A}(2k-1,8) = -v_A(k)\cos(\theta(k))\cos(\phi(k)) \\
	&\bm{A}(2k-1,9) = -w_A(k)\cos(\theta(k))\cos(\phi(k)) \\
	&\bm{A}(2k-1,10) = \sin(\phi(k)) \\
	&\bm{A}(2k-1,11) = 0 \\
	&\bm{A}(2k-1,12) = -\cos(\theta(k))\cos(\phi(k)) \\ \\
	&\bm{A}(2k,1) = 0 \\
	&\bm{A}(2k,2) = 0 \\
	&\bm{A}(2k,3) = 0 \\
	&\bm{A}(2k,4) = u_A(k) \sin(\phi(k)) \\
	&\bm{A}(2k,5) = v_A(k) \sin(\phi(k)) \\
	&\bm{A}(2k,6) = w_A(k) \sin(\phi(k)) \\
	&\bm{A}(2k,7) = -u_A(k)\sin(\theta(k))\cos(\phi(k)) \\
	&\bm{A}(2k,8) = -v_A(k)\sin(\theta(k))\cos(\phi(k)) \\
	&\bm{A}(2k,9) = -w_A(k)\sin(\theta(k))\cos(\phi(k)) \\
	&\bm{A}(2k,10) = 0 \\
	&\bm{A}(2k,11) = \sin(\phi(k)) \\
	&\bm{A}(2k,12) = -\sin(\theta(k))\cos(\phi(k))
	\end{align*}
	
	The vector $\bm{b}$ is defined as follows:
	
	\begin{align*}
	\bm{b}(2k-1) = & - \cos(\theta(k))\cos(\phi(k))z_B(k) \\
	& + \sin(\phi(k))x_B(k) \\ \\
	\bm{b}(2k) = & - \sin(\theta(k))\cos(\phi(k))z_B(k) \\
	& + \sin(\phi(k))y_B(k)
	\end{align*}

\end{document}